\documentclass[10pt,journal]{IEEEtran}
\usepackage{xcolor}
\usepackage{graphicx}
\usepackage{cite}
\usepackage{amsmath,amssymb}
\usepackage{graphicx}
\usepackage[english]{babel}
\usepackage{hhline}
\usepackage{epstopdf}
\usepackage{cite}
\usepackage{siunitx}
\usepackage{amssymb}

\usepackage{amsthm}
\usepackage{mathtools}
\usepackage{siunitx}
\usepackage{amsmath}

\DeclareSIUnit[inter-unit-product={}] \MVA {\mega\volt\ampere} %apparent power
\DeclareSIUnit[inter-unit-product={}] \MWh {\MW\hour} %apparent power
\DeclareSIUnit \pu {p.u.}
\DeclareSIUnit[inter-unit-product={}] \KV {\kilo\volt} %KV

\newtheorem{Remark}{Remark}
\DeclareSIUnit[inter-unit-product={}] \MVA {\mega\volt\ampere} %apparent power
\DeclareSIUnit[inter-unit-product={}] \MWh {\MW\hour} %apparent power
\DeclareSIUnit \pu {p.u.}
\DeclareSIUnit[inter-unit-product={}] \KV {\kilo\volt} 
\renewcommand{\baselinestretch}{0.93}
\include{pythonlisting}
\usepackage{pgfplots}
\usepackage{tikzscale}
\usepackage{float}

\usepackage[export]{adjustbox}
\pgfplotsset{compat=1.12}
\usepackage{caption,subcaption}
\usetikzlibrary{matrix,positioning}
\usetikzlibrary{fit,intersections,}
\makeatletter
\tikzset{ % fitting node, see http://goo.gl/KOvpQ
	fitting node/.style={
		inner sep=0pt,
		fill=none,
		draw=none,
		reset transform,
		fit={(\pgf@pathminx,\pgf@pathminy) (\pgf@pathmaxx,\pgf@pathmaxy)}
	},
	reset transform/.code={\pgftransformreset}
}
\makeatother
\usepackage{filecontents}
\usepackage{nomencl}
\makenomenclature
\renewcommand{\baselinestretch}{0.9}
\begin{document}
	
	\captionsetup[figure]{labelfont={bf},labelformat={default},labelsep=period,name={Fig.}}
	\title{Explainable Incipient Fault Detection Systems for Photovoltaic Panels}
	
	\author{Seshapalli Sairam, Seshadhri Srinivasan~\IEEEmembership{Sr. Member,~IEEE}, Giancarlo Marafioti~\IEEEmembership{Sr. Member,~IEEE}, B. Subathra, Geir Mathisen ~\IEEEmembership{Sr. Member,~IEEE}, Korkut Bekiroglu~\IEEEmembership{Member,~IEEE}
	
	% <-this % stops a space
	%\thanks{Mainak Dan is with the Interdisciplinary
	%Graduate School, Nanyang Technological University, Singapore
	%e-mail: MAINAK001@e.ntu.edu.sg}% <-this % stops a space
	\thanks{Seshapalli Sairam is with Dept. of Instrumentation and Control Engineering,
Kalasalingam Academy for Research and Education, Krishnankoil,
Srivilliputtur, India.
 	e-mail : s.sairam@klu.ac.in}
	\thanks{Seshadhri Srinivasan is with GE Corporate Research Center, Bangalore-560068.
	e-mail : seshadhri.srinivasan@ge.com}% <-this % stops a space
		\thanks{Giancarlo Marafioti is with SINTEF, Cybernetics.
	e-mail : giancarlo.marafioti@sintef.no}% <-this % stops a space
		\thanks{B. Subathra is with Dept. of Instrumentation and Control Engineering,
Kalasalingam Academy for Research and Education, Krishnankoil,
Srivilliputtur, India.
 	e-mail : b.subathra@klu.ac.in}% <-this % stops a space
		\thanks{Geir Mathisen is with Norwegian University of Science and Technology Faculty of
Engineering, Cybernetics.
 	e-mail : geir.mathisen@ntnu.no}
 	\thanks{K. Bekiroglu is with the College of Engineering, SUNY Polytechnic Institute, Utica 13503, NY, USA.
 	e-mail : korkut.bekiroglu@sunypoly.edu}}% <-this % stops a space
%	\thanks{Suresh Sundaram is with the School of Computer Science and Engineering, Nanyang Technological University, Singapore. e-mail : ssundaram@ntu.edu.sg}% <-this % stops a space
%	\thanks{Luigi Glielmo is with the Department of Computer Engineering, Università degli Studi del Sannio, Italy. e-mail : glielmo@unisannio.it}}% <-this % stops a space

	%\markboth{IEEE Transactions on Sustainable Energy,~Vol.~XXX, No.~X, Oct ~2018}%
	%{Shell \MakeLowercase{\textit{et al.}}: Bare Demo of IEEEtran.cls for IEEE Journals}i
	\maketitle
	\begin{abstract}
%Detecting incipient faults in photovoltaic (PV) panels is a challenging task, and the problem accentuates at low irradiance conditions. 

%The incipient faults in photo-voltaic (PV) panels are characterized by intermittent behaviours and signal thresholds which are very hard to detect. 

This paper presents an eXplainable Fault Detection and Diagnosis System (XFDDS) for incipient faults in PV panels.  The XFDDS is a hybrid approach that combines the model-based and data-driven framework. Model-based FDD for PV panels lacks high fidelity models at low irradiance conditions for detecting incipient faults. To overcome this, a novel irradiance based three diode model (IB3DM) is proposed. It is a nine parameter model that provides higher accuracy even at low irradiance conditions, an important aspect for detecting incipient faults from noise.  To exploit PV data, an  extreme gradient boosting (XGBoost) is used due to its ability to detecting incipient faults. Lack of explainability, feature variability for sample instances, and false alarms are challenges with data-driven FDD methods. 
These shortcomings are overcome by hybridization of XGBoost and IB3DM, and using eXplainable Artificial Intelligence (XAI) techniques. To combine the XGBoost and IB3DM, a fault-signature metric is proposed that helps reducing false alarms and also trigger explanation on detecting incipient faults. To provide explainability, an eXplainable Artificial Intelligence (XAI) application is developed.  It uses the local interpretable model-agnostic explanations (LIME) framework and provides explanations on classifier outputs for data instance.  These explanations help field engineers/technicians for performing troubleshooting and maintenance operations. The proposed XFDDS is illustrated using experiments on different PV technologies and our results demonstrate the perceived benefits.

	\end{abstract}

	\vspace{-0.5cm}
			\mbox{}
\setlength{\nomitemsep}{-\parsep}
\nomenclature{$I_p$ }{Current generated by incident light [A];}
\nomenclature{{$I_{s1}$},~{$I_{s2}$},~{$I_{s3}$}}{Diode current [A];}
\nomenclature{{$I_{l1}$}~,{$I_{l2}$}~,{$I_{l3}$} }{Reverse saturation or leakage current [A];}
\nomenclature{$I_{sh}$}{Shunt current [A];}
\nomenclature{{$n_{01}$}, {$n_{02}$},{ $n_{03}$}}{Diode ideality factor;}
\nomenclature{$R_{s}$ }{Series resistance [\si{\ohm}];}
\nomenclature{$R_{sh}$}{Parallel resistance [\si{\ohm}];}
\nomenclature{$I$}{Solar cell output current [A];}
\nomenclature{$V$}{Solar cell output voltage [V];}
\nomenclature{$P$}{Power [\si{\watt}];}
\nomenclature{$STC$}{Standard Test Condition [0-1000 \si{\watt\per\square\meter}, 25\si{\degree}C];}
\nomenclature{$N_s$ }{Number of solar cells connected in series;}
\nomenclature{$G$}{Irradiance on device surface [\si{\watt\per\square\meter}];}
\nomenclature{$G_{STC}$}{Standard irradiance on device surface [\si{\watt\per\square\meter}];}
\nomenclature{$I_{p,n}$}{Photo-generated current at STC [A];}
\nomenclature{{$I_{SC, STC}$}}{Short circuit current at STC [A];}
\nomenclature{$I-V$}{Current-Voltage characteristics;}
\nomenclature{$V_{OC, STC}$}{Open circuit voltage at STC [V];}
\nomenclature{{$T_r$,~$T_{STC}$}}{Actual and STC temperature [\si{\degree}C];}
\nomenclature{$k$}{Boltzmann constant [$1.3806*10^{-23}$ \si{\joule\per\kelvin}];}
\nomenclature{$q$}{Electron charge [$1.6022*10^{-19}$ C];}
\nomenclature{$V_t$}{Thermal voltage [$V$];}
\nomenclature{$K_I$}{Temperature coefficient of the short–circuit current [A/\si{\degree}C];}
\nomenclature{$K_V$}{Temperature coefficient of the open–circuit voltage [V/\si{\degree}C];}

\printnomenclature[0.79in]

\vspace{0.1cm}
%\begin{table}[h]
%\caption{ Abbrivations \label{tab2}}
%\centering
%\begin{tabular}{  c  l }
%\hline
%Abbrivations & Description  \\
%\hline
%SDM& single diode model\\
%DDM& double diode model\\
%CD3DM& current dependent three diode moodel\\
%IB3DM& irradiance based three diode model\\
%PSO& particle swarm optimization\\
%TLBO& teaching-learning based optimization\\
%BBO&biogeography-based optimization\\
%SFLA& shuffled frog leaping algorithm\\
%    RMSE& root mean square error \\
%\hline
%\end{tabular}
%\end{table}
	\IEEEpeerreviewmaketitle

	\vspace{-0.1cm}
	
	%\section*{Acronym}
	%\begin{table}[h]
	%\begin{tabular}{ll}
	%MES & Multi-energy System\\
	%HVAC & Heating, Ventilation and Air-conditioning \\ & System\\
	%PV & Photo-voltaic Panel\\
	%GT & Gas Turbine\\
	%DG & Generator Unit\\
	%ESS & Electrical Storage System\\
	%TES & Thermal Energy storage\\
	%CHP & Chiller Plant\\
	%Ab & Absorption Chiller\\
	%McFIS & Meta-cognitive Fuzzy inference System\\
	
	%MI(N)LP & Mixed-integer (non)linear programming\\
	%RCGA & Real-Coded Genetic Algorithm\\

	%\end{tabular}
	%\end{table}
	\IEEEpeerreviewmaketitle
	\section{Introduction}
\subsection{Motivation}	
\IEEEPARstart{E}{xponential} growth in photovoltaic (PV) deployments has raised interest in its reliable operation  ~\cite{rajesh2020design}. As PV panels are installed in harsh environments and subjected to varying weather conditions,  they are prone to diverse faults (permanent, incipient, and intermittent) with different severity levels~\cite{karmakar2020detection}. Such faults could diminish energy production, accelerate aging, and even cause fire hazards~\cite{zhao2014graph}. Therefore, detecting and locating faults early (at the incipient stage)  is pivotal for the PV panel's reliable operations~\cite{pillai2019comparative}. Detecting incipient faults challenging as the signatures are less evident due to low magnitude, and  the problem accentuates at low irradiance conditions.  Also, incipient faults are quite intermittent and show up for a short duration. Therefore, detecting them becomes more challenging.  Nevertheless, incipient faults could develop as a severe fault in the long-run if left undetected/unattended, leading to costly replacements and maintenance operations ~\cite{jin2019}.  Consequently, detecting incipient faults has gained significant traction recently~\cite{garoudja2017statistical}. While fault-detection and diagnosis (FDD) systems are proven to improve PV system reliability~\cite{mellit2018fault}, there are few challenges that need to be addressed for detecting incipient faults: {\em{(i)}} high fidelity PV models providing good accuracy at low irradiance conditions are required, {\em{(ii)}} existing FDD methods cannot reason their decisions to the field engineers/technicians, and {\em{(iii)}} difficult distinguishing between false alarms and incipient faults. Our objective in this paper is to propose an eXplainable Fault Detection and Diagnosis Systems (XFDDS) for incipient faults in PV panels that address challenges with FDD systems.
\vspace{-0.3cm}
\subsection{Literature Review}
The FDD methods in the literature can be broadly discerned as being--- model-based (MB), signal-based (SB), and data-driven (DD)~\cite{gao2015}. The MB methods use PV panel models (set of nonlinear equations) followed by signal-analysis (e.g., correlation analysis) on the input-output data from the model to detect faults~\cite{chaibi2019simple}. The single-diode model (SDM)~\cite{shongwe2015comparative}, double-diode model (DDM)~\cite{bradaschia2019parameter}, and three-diode models (TDM)~\cite{khanna2015three,qais2019identification} are widely used in FDD systems. While existing models provide good accuracy at high irradiance, their accuracy is less at low irradiance conditions. The SB methods use fault signatures from sensor data to detect faults ~\cite{triki2018fault}. Widely used SB methods are: statistical signal processing ~\cite{davarifar2013real}, I-V (current-voltage) characteristics analysis~\cite{fadhel2019pv}, power loss analysis~\cite{ali2017real} and, voltage and current measurements~\cite{chen2017adaptive}. More recently, SB-FDD methods using two-stage support vector machines~\cite{yi2017line}, multi-signal decomposition, and fuzzy inference systems~\cite{yi2016fault} have also been proposed. The DD methods using labeled fault-data and artificial intelligence (AI) techniques have shown promise in improving detection accuracy due to their powerful model representation capabilities.
 The DD methods leverage historical labeled data and powerful models from AI techniques to perform multi-class regression or classification, which are quite important for detecting faults~\cite{zhao2019artificial}. In the literature, FDD methods using AI models such as  random forest~\cite{dhibi2020reduced}, collaborative filtering~\cite{zhao2020collaborative}, extreme gradient boosting ~\cite{gan2019novel}, and  such techniques have been proposing~(see,\cite{ebrahimifakhar2020data} and references therein).  Despite these advances, detecting incipient faults addressing the fundamental challenges of accuracy at low irradiance conditions, lack of explainability on decisions to field engineers/technicians, and distinguishing false alarms from incipient faults is rather unexplored in the literature to our best knowledge.
\vspace{-0.3cm}
 \subsection{Contributions}
  This paper proposes an XFDSS  for PV panels addressing challenges with FDD system. It is a hybrid method that combines model-based and data-driven approaches. To overcome accuracy challenges under low irradiance conditions and detect incipient faults, an irradiance based three diode model (IB3DM) proposed. The model uses irradiance and temperature in parameter computations inherently, thereby increasing its accuracy even at low irradiance conditions.  For fault explainability and distinguishing false alarms from incipient faults, the IB3DM is combined with data-based approaches that perform multi-label classification. This paper uses the extreme gradient boosting (XGBoost) based multi-label classifier due to its suitability to detect incipient faults. As XGBoost cannot explain its decisions to the field technician/engineer, recently developed eXplainable AI (XAI) techniques are used.  The XAI extends the capabilities of the AI techniques by providing explanations on decisions on individual data-instances~\cite{li2020survey}, a key aspect in incipient fault detection. We show that these explanations are very useful for field engineers/technicians to understand the fault-causes and fault-type.  The local interpretable model-agnostic explanations (LIME) approach is used~\cite{bramhall2020qlime} to provide the explanations. The main idea is to perturb the features and compute the importance and variable thresholds for being classified as faults on individual samples. Main contributions are: 
 %\vspace{-0.1cm}
\begin{enumerate}
	\item[1.] A novel three diode model called the Irradiance Based Three Diode Model (IB3DM) which inherently captures the influences of solar irradiance and ambient temperature;  
	\item[2.]  Design an XFDS leveraging the accuracy of IB3DM,  XGBoost, and LIME; 
	\item[3.] Illustrate the IB3DM and XFDS using experiments and simulations on different PV technologies.
	%\item [4] A methodology for smartly detecting partial shading conditions using the proposed model with meteorological forecasts is also presented.
\end{enumerate}

The paper is organized into five sections. The components of XFDDS is explained in Section II. The IB3DM for PV panel and its parameter computation is explained in Section III. The XFDDS methodology is presented in Section IV. Results are presented in Section V and conclusions are presented in Section VI.

\section{Explainable Fault Detection and Diagnosis System}
The main challenges with existing FDD techniques are:
\begin{itemize}
\item[{\em{(C1)}}] Lack of high fidelity models capturing PV panel performance at low irradiance conditions;
\item[{\em{(C2)}}] Existing FDD methods lack explanations to field engineers/technicians on why a particular sample was classified as faults and the variable thresholds on which this decision on a fault is made; 
\item[{\em{(C3)}}] Data-based models compute feature importance for a particular fault on the global data, whereas incipient faults are intermittent and there are inconsistencies within data instances as well;
\item[{\em{(C4)}}] Data-based models cause false alarms due to mis-classification.

\end{itemize}

		\begin{figure}[!htb]
		%\centering
			\centering
		\includegraphics[height=5.0cm, width=8.2cm]{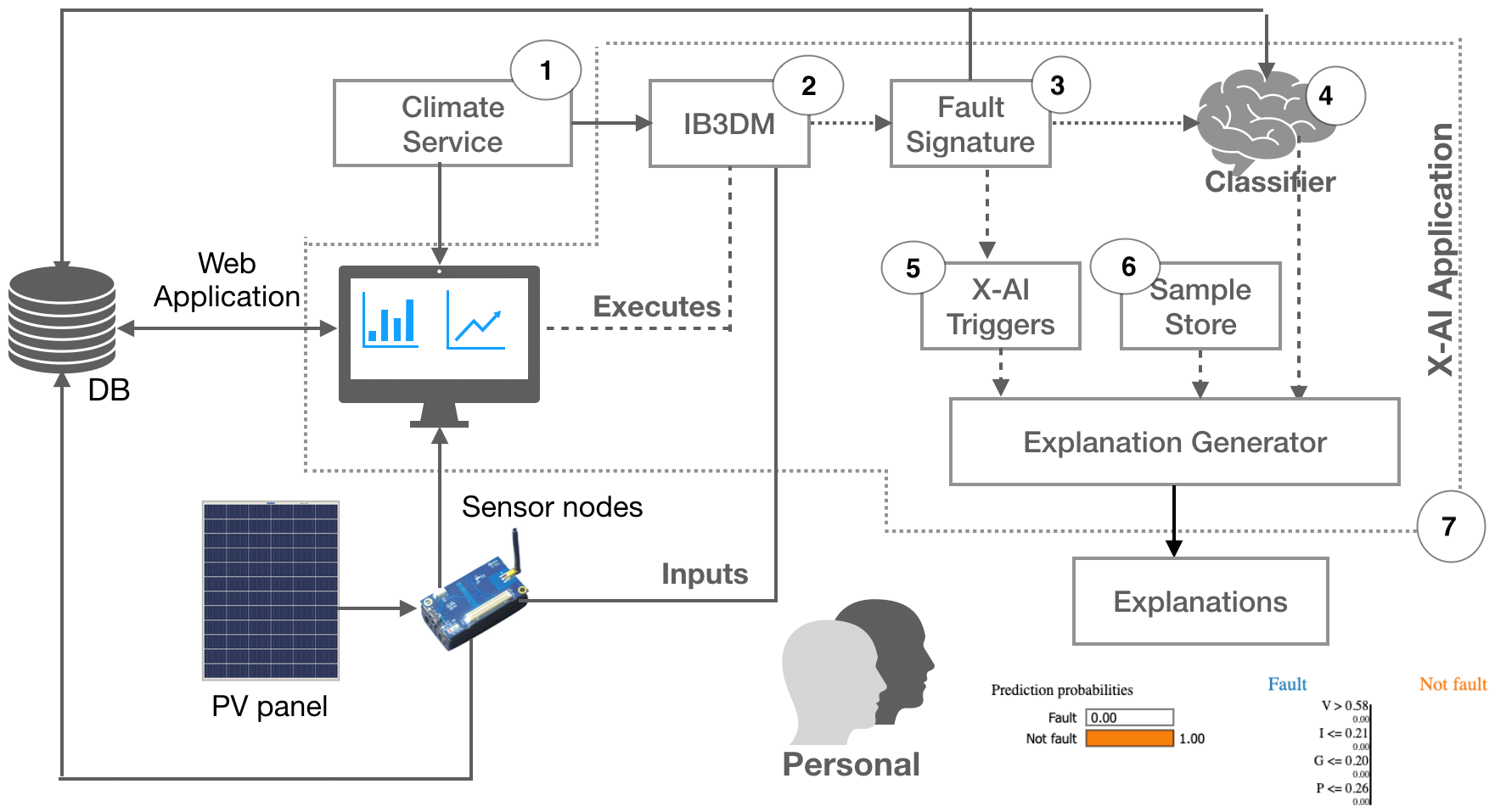}
				\caption{Explainable fault detection system}
		\label{fig:Incipient}
	\end{figure}

The XFDDS proposed in our work addresses the challenges (C1)-(C4) and its schematic is shown in  Fig.~\ref{fig:Incipient}. Its main components are: climate service, IB3DM, fault-signature metric, the XGBoost classifier, XAI triggers, sample store and XAI application. The IB3DM uses the climate service (a web-application) to obtain solar irradiance and temperature for predicting the PV outputs  (voltage, current, and power).
Exploiting IB3DMs' model accuracy, a fault-signature metric is defined (see Section \ref{sec:metric}) which serves as a trigger for obtaining explanations from XAI application and such samples are stored in {\it{sample store}}, a local cache. The XGBoost based classifier is a combination of multiple classifiers, and regression tree (CART) ensemble created using boosting techniques~\cite{gan2019novel}. The XGBoost is selected as the data-based model in our application, as it naturally fits the incipient fault-detection framework as detailed later. Two challenges with XGBoost are lack of explainability and false alarms~\cite{torres2019regression}. Moreover, the feature importances for a particular fault are computed based on global data, contrary to this incipient faults are intermittent with data varying among fault samples (feature inconsistency problem). The particular of XAI application generate these explanations on a data-instances, which is very important for incipient faults. These explanations help the user to identify the fault types and variable thresholds based on which the fault was detected. In what follows the IB3DM model is first proposed and then the XFDDS approach is illustrated.

	\section{Novel Irradiance-based three diode model}
	\label{sec:IB3DM}

%This section proposes a fault detection method is explained. Two main components of eXplainable Fault Detection and Diagnosis System (XFDSS) are Irradiance-based three diode model (IB3DM) and eXplainable AI (XAI) application. This section presents the IB3DM for PV panels, its equivalent circuit and model equations. After defining the IB3DM model, its model parameters needs to be identfied to use in XAI application. Therefore, an optimization problem is formulated to calculate these parameters. Further, parameter computations of IB3DM is shown to be non-convex and non-linear optimization problem. These optimization and its solution technics are presented in this section and the parameter computations are given in the results section.
%\vspace{-0.5cm}
%\subsection{Irradiance Based Three Diode Model}

Our model parameters depend on irradiance and module temperature and it addresses the challenge (C1). Therefore, we call our model irradiance-based three diode model (IB3DM).  The IB3DM is an extension of the TDM proposed in~\cite{khanna2015three,qais2019identification} wherein $I_{P}$, the light generated current depends on irradiance and module temperature. Further, in IB3DM, the ideality factors are not fixed; rather, they are obtained as a solution to an optimization problem by specifying bounds ([0,~2]). This is a deviation from existing works in three diode models where higher ideality factors are used leading to low fill-factor that could be achieved only in industrial-grade panels. This makes existing TDM unsuitable for residential PV panels. %The equivalent circuit, model equations, and parameter computations are explained in this next section. 

	\vspace{-0.5cm}
	\subsection{Equivalent Circuit and Model Parameters}
	\label{subsec:IB3DM}
\begin{figure}
		\centering
		\includegraphics[height=3.0cm, width=8cm]{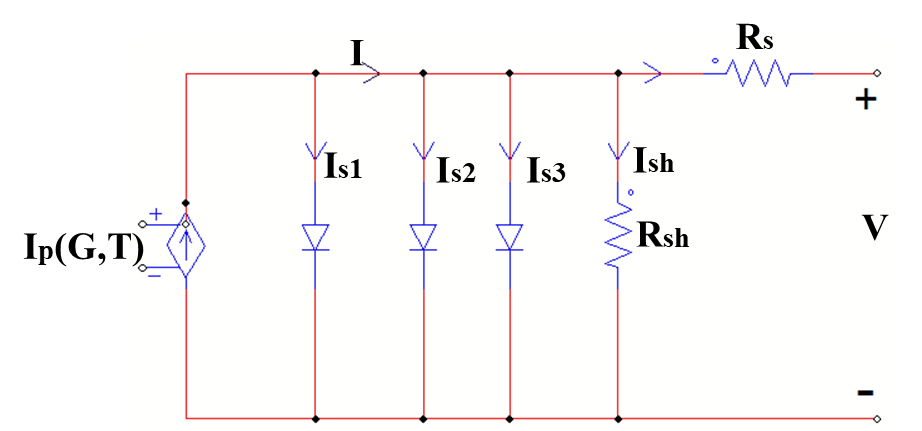}
		\caption{Equivalent circuit of IB3DM.}
		\label{fig:4}
	\end{figure}

Our idea is to propose a three diode model that accurately captures the PV cells’ performance even under low irradiance conditions. 
\begin{Remark} In IB3DM, the source current is modelled as a dependent current source  that is a function of irradiance and module temperature, denoted by $I_{p}(G,~T)$. The photo-generated current has two parts; the first part which is a premultiplier is linearly dependent on the irradiance and acts as a scaling factor for the second part that depends on panel temperature.
\end{Remark}
Suppose the nominal phase current is denoted by $I_{p,n}$, then the dependence on phase current on the irradiance is given by,
\vspace{-0.1cm}	
\begin{equation}\label{eq:5}
	I_{P}(G,T)= \Bigg(\frac{G}{G_{STC}}\Bigg) \Big(I_{p,n}+K_{I}\times(T-T_{r})\Big)
	\end{equation}
	
\noindent where $K_I$ is a constant computed from data-sheets. The source current is in parallel to three diodes with a series resistance $R_s$ and shunt resistance $R_{sh}$, as shown in Fig.~\ref{fig:4}. With this modification, the current source is a function of irradiance and module temperature. From the equivalent circuit, the output current is given by

	{\small{
	\begin{equation}\label{eq:6}
	I=I_{P}-I_{l1}-I_{l2}-I_{l3}-I_{sh}
	\end{equation}}}
Similarly, current through the shunt resistance $R_{sh}$ in Fig.~\ref{fig:4} is,

{\small{
	\begin{equation}\label{eq:7}
	I_{sh}=\frac{V+IR_{s}}{R_{sh}}
	\end{equation}
	}}
	
The diode saturation currents can be calculated as:

	\begin{equation}\label{eq:8}
	I_{01}=\frac{I_{S}+K_{I}\times(T-T_{r})}{exp\big(\frac{V_{OC}+K_{V}\times(T-T_{r})}{{V_{t}\times n_{01}}}\big)-1},
	\end{equation}

		\begin{equation}\label{eq:62}
		I_{02}=\frac{I_{S}+K_{I}\times(T-T_{r})}{exp\big( \frac{V_{OC}+K_{V}\times(T-T_{r})}{{V_{t}\times n_{02}}}\big)-1},
		\end{equation}

		\begin{equation}\label{eq:67}
		I_{03}=\frac{I_{S}+K_{I}\times(T-T_{r})}{exp\big(\frac{V_{OC}+K_{V}\times(T-T_{r})}{{V_{t}\times n_{03}}}\big)-1},
		\end{equation}
		
 The saturation currents strongly depends on the temperature as indicated by equations \eqref{eq:8}-\eqref{eq:67}. Note that the coefficient $K_{V}$ is from the manufacturers' data-sheet and used to compute I-V curve for different temperatures as seen the equations \eqref{eq:8}-\eqref{eq:67}. The junction thermal voltage is given by,
   \begin{equation}\label{eq:9}
	V_{t}=\frac{N_{s}kT}{q}.
	\end{equation}
Combining equations \eqref{eq:5}-\eqref{eq:9}, one can obtain the equations relating the output current, output voltage, and model parameters for the IB3DM:
	\begin{equation}\label{eq:10}
	\begin{split}
	I
	&= I_{P}-I_{01}\Bigg(exp\Big({\frac{V+IR_{s}}{n_{01}V_{t}}}\Big)-1\Bigg)\\
	&
	-I_{02}\Bigg(exp\Big({\frac{V+IR_{s}}{n_{02}V_{t}}}\Big)-1\Bigg)\\
	&  -I_{03}\Bigg(exp\Big({\frac{V+IR_{s}}{n_{03}V_{t}}}\Big)-1\Bigg)-\frac{V+IR_{s}}{R_{sh}},
	\end{split}
	\end{equation}	

\noindent where $n_{01}$, $n_{02}$, and $n_{03}$ are diode ideality factors. Consequently, the IB3DM has nine parameters given by $\mathcal{P} _{IB3DM}= \{I_P, I_{l1}, I_{l2}, I_{I3}, R_S, R_{sh}, n_{01},n_{02},n_{03}\}$ which should be obtained from I-V curve data. Next step is to compute the model parameters and an optimization based approach is proposed as detailed in the next section.

%These parameters are computed by solving a nonlinear system of equations that is computational complex to solve and obtain a solution. To overcome this computational complexity, optimization approaches are used for parameter estimation. However, the parameter identification model lends itself to a nonlinear and non-convex problem, and therefore, meta-heuristic approaches are employed to search the solution. In the following section, the objective functions and the system of equations of the optimization problem that should be solved using meta-heuristic approaches is explained.

  		\vspace{-0.5cm}
	
	\subsection{The IB3DM Parameter Computation}
To compute the IB3DMs' model parameters, first the $I_{P}$ is calculated using $G$ and $T$ from equation \eqref{eq:5}. Next, we define the objective function as the root mean square error between the experimental V-I and the estimated model is given by,
\vspace{-0.1cm}			\begin{equation}
	\begin{split}
		f_{m}(V,I,\mathcal{P})
	&= I - I_{P}+I_{01}\Bigg(exp\Big({\frac{V+IR_{s}}{n_{01}V_{t}}}\Big)-1\Bigg)\\
	&
	+I_{02}\Bigg(exp\Big({\frac{V+IR_{s}}{n_{02}V_{t}}}\Big)-1\Bigg)\\
	&  + I_{03}\Bigg(exp\Big({\frac{V+IR_{s}}{n_{03}V_{t}}}\Big)-1\Bigg)+\frac{V+IR_{s}}{R_{sh}}
	\end{split}
	\end{equation}

\noindent It measures the difference between experimental I-V curve data and the one calculated using the model, i.e., ${f_{m}(V,I,\mathcal{P})}=I_{measured}-I_{calculated}$ with $V$ varying over the operating range. To compute the model parameters, we utilized the root mean square error (RMSE) as a metric denoted by, $\mathcal{J}=\sqrt{\frac{1}{N_{e}}\sum_{i=1}^{N_{e}} f_{m}(V,I,\mathcal{P})^2}$.
The parameter computation problem is modelled as an optimization problem given by,
\begin{align}\label{eq:14}
&\underset{\mathcal{P}_{IB3DM}}{\operatorname{min}}~~~~~~~~~ \mathcal{J}\\ \nonumber
	s.t.&~~~~~~~~~~~~~~~~~\eqref{eq:6}-\eqref{eq:10} \\ \nonumber	
%	&w(G) = \frac{1}{e^{(-\frac{\|G-G_s\|_2}{\Sigma})^2}}\\ \nonumber
%	&I_{P}(G,T_r)= \Bigg(\frac{G}{G_s}\Bigg) I_{p,n}+K_{I}*(T-T_{r})\\ \nonumber
%	& I=I_{P}-I_{s1}-I_{s2}-I_{s3}-I_{sh}\\ \nonumber
%	& I_{sh}=\frac{V+IR_{s}}{R_{sh}} \\ \nonumber
%	& 	I_{0,n}=\frac{I_{S}+K_{I}*(T-T_{r})}{exp\big(\frac{V_{OC}+K_{V}*(T-T_{r})}{{V_{t}\sum_{i=1}^{l}n_{i}}}-1\big)}, \\ \nonumber
%	&V_{t}=\frac{N_{s}kT}{q}, \\ \nonumber
%	&I= I_{P}-I_{l1}\Bigg(exp\Bigg({\frac{V+IR_{s}}{n_{01}V_{t}}}\Bigg)-1\Bigg)\\ \nonumber
%	&~~~~~ -I_{l2}\Bigg(exp({\frac{V+IR_{s}}{n_{02}V_{t}}}\Bigg)-1\Bigg)\\ \nonumber
%	&~~~~~~~~~~  -I_{l3}\Bigg(exp\Bigg({\frac{V+IR_{s}}{n_{03}V_{t}}}\Bigg)-1\Bigg)-\frac{V+IR_{s}}{R_{sh}}.\\\nonumber
\end{align}
\vspace{-0.5cm}

	Clearly, \eqref{eq:14}  is nonlinear and non-convex that is computationally complex to solve using conventional optimization techniques. Usually meta-heuristic approaches are used for computations for this type of optimization problems~\cite{chin2015cell}. Our analysis utilizes five different meta-heuristic algorithms that are explained in results section.

\section{Explainable Fault Detection and Diagnosis Methodology}

%\noindent {\bf{Note:}} The LIME approach is model-agnostic, i.e., the approach explains the model without peeking inside it. While XGBoost is used in our paper due to its suitability to detect incipient faults, any such decision tree algorithms could be used.

\vspace{-0.1cm}
\subsection{ Extreme Gradient Boosting for Fault Classification}
Gradient boosting decision tree is a powerful machine learning algorithm wherein multiple weak learner ensemble form a strong learner~\cite{sun2020gradient}. The XGBoost main idea is that the training set in the current instance is related to the learning results from previous learning, and the weights on data-samples are adjusted on each sample in each iteration. These features naturally fits the incipient fault-detection scenario as labeled fault-data are available sequentially, and weight adaptation in each sample increases model accuracy. Furthermore, in XGBoost the current decision tree (leaf) is fitted based on the residuals from the previous trees and it uses the gradient boosting principle  wherein new decision trees are constructed to correlate to the negative gradient of the loss-function. A detailed description of XGBoost could be found in~\cite{fan2018comparison,sapountzoglou2020fault}.

\emph{This investigation uses XGBoost for detecting two incipient faults: {\em{(i)}} line-to-line (LL) fault  which denotes short-circuit within the string or on multiple PV strings  and {\em{(ii)}} partial shading.} Two different classification cases are considered: binary and multi-class classification. In binary classification, labeled data about whether a given sample is faulty or healthy operating condition is predicted without considering the fault-type (LL or partial shading), whereas in multi-class fault classification, the fault-type are also included as additional class.   
\begin{figure}[h]
\centering
\includegraphics[height=4.5cm, width=8cm]{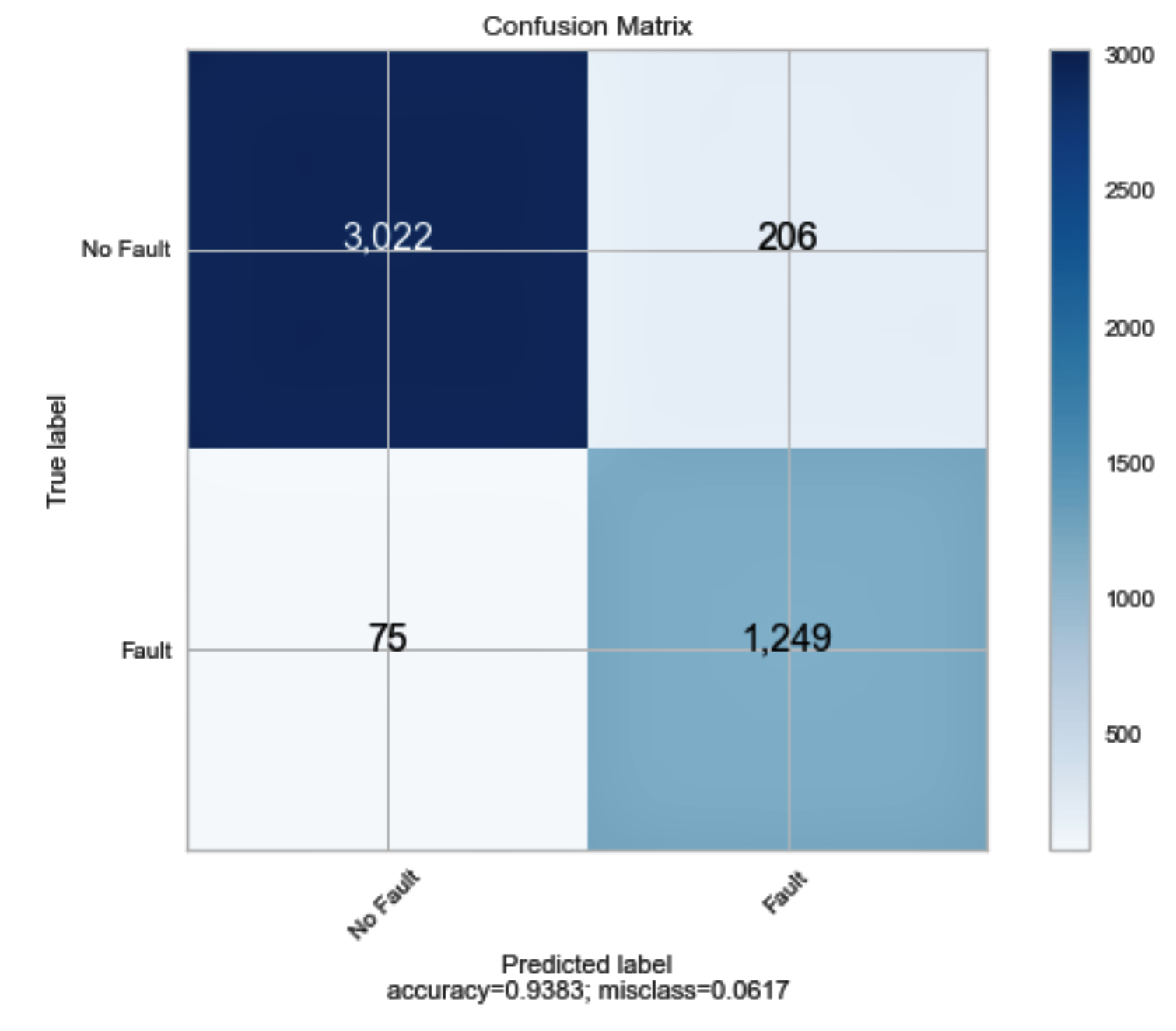}
\caption{Confusion Matrix for Binary Classification}
\label{fig:CF1}
\end{figure}

In binary classification, voltage, current, irradiance, and power are the inputs and a binary variable modelling the sample to be faulty/healthy is the output. Dichotomous search optimization was used for tuning hyper-parameters that resulted in 100 learners.  During training, the classifier showed 93-95\% accuracy, and validation had 86-93\% accuracy.  The confusion matrix for the binary classifier is shown in Fig.~\ref{fig:CF1}. One can see that the classifier performs extremely well for binary classification.

\begin{figure}[h]
\centering
\includegraphics[height=5cm, width=8cm]{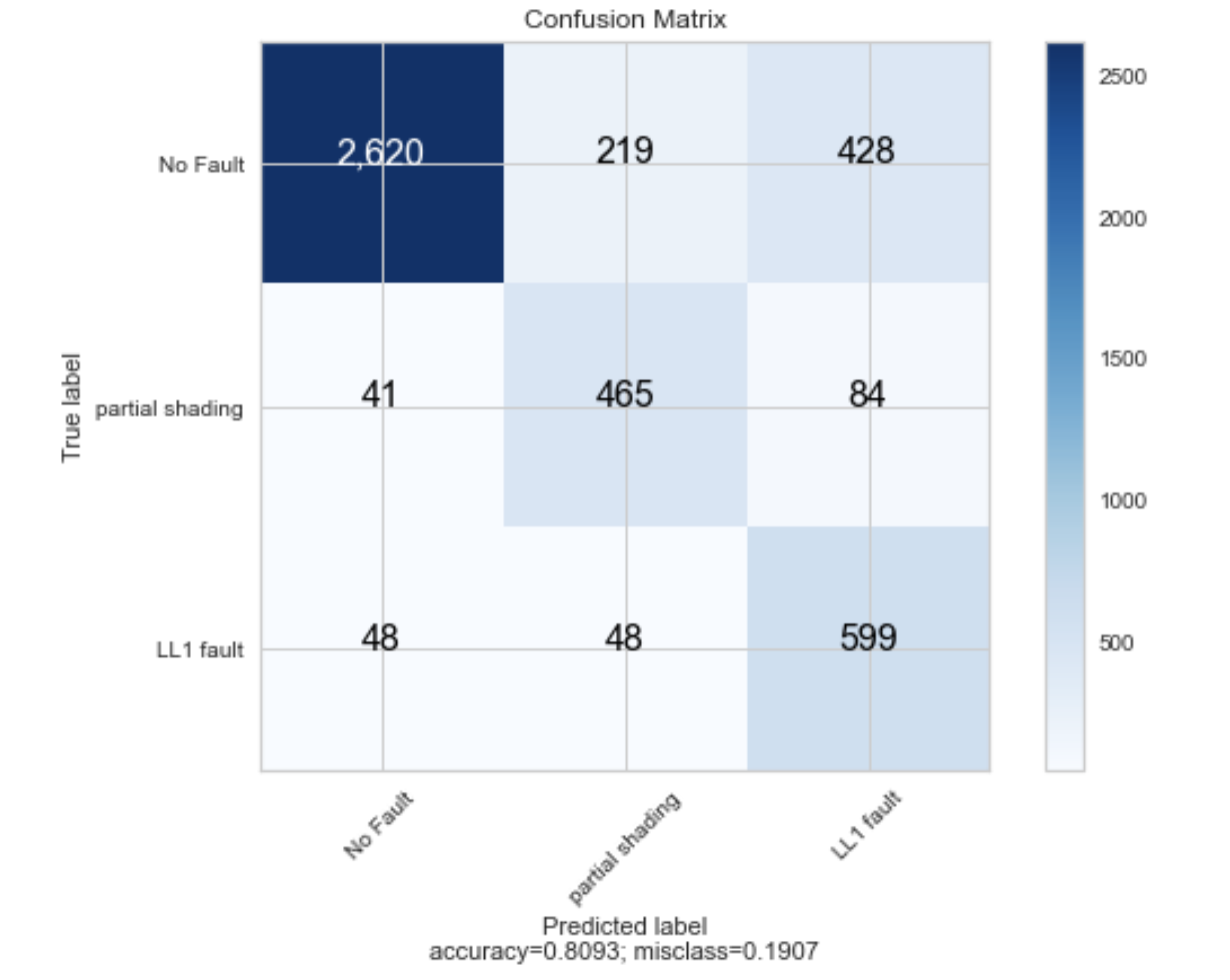}
\caption{Confusion Matrix for Multi-Class Classification}
\label{fig:CF2}
\end{figure}

In multi-class classification, additional labels on the fault type (LL), or partial shading) is added to the data-set. Then the XGBoost classifier is trained to identify the fault label. The XGBoost hyper parameters was optimized using dichotomous search and it had 130 learners. During the training, the classifier showed 88-92\% accuracy, and 78-82\% during validation. Confusion matrix for the multi-class classification during validation is shown in Fig.~\ref{fig:CF2}. While its accuracy is reasonable, the challenges (C2)-(C4) are not addressed by XGBoost.

%One can see that the classifier can provide good accuracy and distinguish between healthy operation/LL faults. However, there are some mis-classification which could trigger false alarms. Additionally, there are no explanations on fault causes to the field technician/engineer. 

\subsection{Explainable Fault Detection} \label{sec:metric}
To provide local explanations and address challenges (C2)-(C4), the data-driven approach is first fused with model-based approach by proposing a fault-signature metric (FSM) given by,
\begin{equation}
\sigma(G,P)= \gamma(G)~exp\big({\frac{-\|G-G_s\|_{\ell_2}
}{\Sigma_G}}\big)^{-1}  \times  exp\big({\frac{-\|P-\hat{P}\|_{\ell_2}
}{\Sigma_P}}\big)^{-1},
\end{equation}
\noindent  where $\gamma(G),\Sigma_G$, and $\Sigma_P$ are the scaling factor as a function of irradiance, variances in the actual solar irradiance, and power generated by the solar panel, respectively. Also, $\|~\|_{\ell_2}$ represents $\ell_2$ norm. $\hat{P}$ is the estimated power computed by the IB3DM, and the signature computes the deviations in power. The fault-signature metric serves two purposes: {\em{(i)}} it weights low variability at low irradiance conditions higher than at high irradiance conditions helps overcoming noise and preventing false alarms and {\em{(ii)}} triggers for explanations on detecting an incipient fault by using thresholds on fault-signature metric.
%, and {\em{(iii)}} helps identifying points wherein the error between the model and actual PV output is quite high. Additionally, fault-signature metric could also be used as additional feature for the classifier to improve its accuracy which is not fully explored in this work.

The XFDDS uses FSM to eliminate false alarms by comparing it with results of the XGBoost. Second, event triggers for fault explanations are generated using FSM. Once the FSM exceeds a known threshold, explanations are asked from the XAI application and such samples are stored in sample store.

On receiving triggers, the XAI application is activated that uses the local interpretable model-agnostic explanations (LIME) ~\cite{bramhall2020qlime} framework. It utilizes surrogate modelling wherein the model is considered a black-box, and the features are perturbed to find feature importance on a particular sample. The data instance is perturbed, and samples are generated from the data-set distribution, which is weighted based on their distances from the current point. Then feature selection is applied to keep the relevant variables, and the linear model is trained on the weighted data-set automatically within the algorithm. Once trained, the model explains to the user about the variables and their thresholds that made the model decide a particular sample instance as a fault/normal operation. This is quite useful in detecting incipient faults as they are intermittent and occur for a few fault-samples.

\emph{While the core of XFDS is still the XGBoost, the XAI extracts explanations on why a particular sample was classified as being faulty/healthy.} Further, thresholds on variables that helped make these decisions are also provided, which is quite useful in detecting even the fault type. For example, a LL fault is characterized by high voltage but a lower current and power. As power drops due to circulating currents are very hard to predict. Moreover, intermittent LL faults are difficult to catch FDDS. However, with XAI could reason out incipient LL faults.

\begin{figure}[h]
	\centering
	\includegraphics[height=4.5cm, width=8cm]{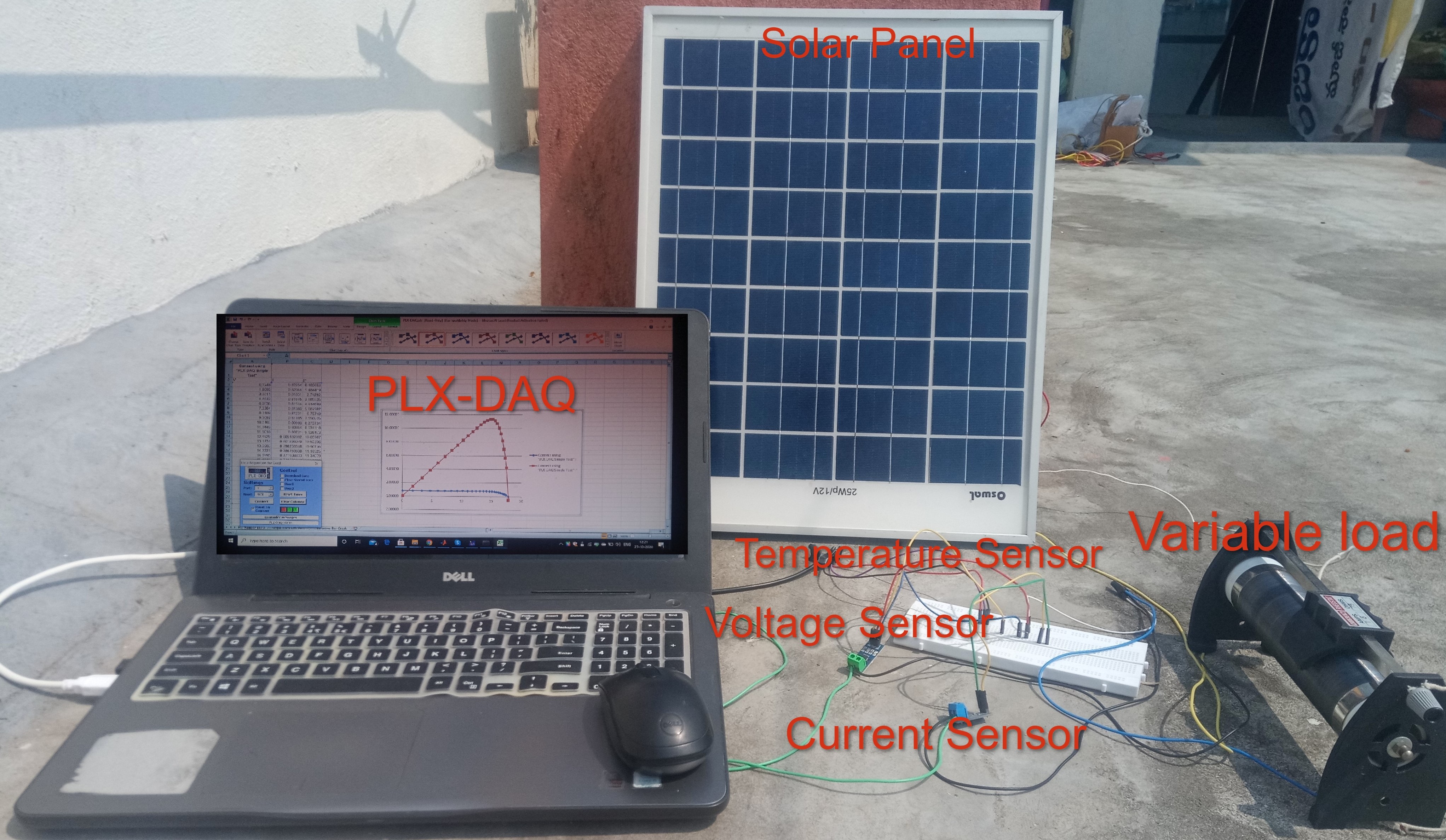}
			\caption{Experimental setup using a PV module.}
	\label{fig:experiment}
\end{figure}

\vspace{-0.3cm}
\section{Results}	
The proposed XFDDS could be implemented on simple hardware as illustrated in this section. In our experiments, the fault-detection is implemented by interfacing embedded hardware with computer. However, the computer could be replaced with any system-on-chip (SoC) with limited computing power. The computer interfaces to the web application directly and obtains the weather data. The weather station recordings are ported as a comma-separated variable. The ATmega 328P processor is used as the data-acquisition unit, which interfaces to the computer through the software application and receives the measurements from sensors. Measurements were obtained from the current sensor (INA169), temperature sensor (DS18B20), and voltage sensor (F031-06), respectively. The data was transmitted to the computer using the UART (Universal Asynchronous Receiver/Transmitter), a serial communication mode to PLX-DAQ and the measurements are stored in a data-base for further processing. Temperature sensor Maxim IC DS 18B20 is a 1-wire digital temperature sensor that reports temperature in Celsius with 9-12 bit precision and has a working range of -55 to 125\si{\celsius}. A rheostat is used as the load, and the experiments are conducted at International Research Center, Kalasalingam University, India. Experiments with PV panel, sensors interfaced with computers, and Python was used to implement the fault-detection scheme (see Fig.~\ref{fig:experiment}).
	
\subsection{The IB3DM Model Parameter Estimation}
As stated earlier, the IB3DM parameter computation requires solving a nonlinear and non-convex optimization problem in \eqref{eq:14}. To overcome computational difficulties, five different meta-heuristic algorithms are used: {\em{(i)}} firefly, {\em{(ii)}} particle swarm optimization (PSO), {\em{(iii)}} teaching-learning based optimization (TLBO), {\em{(iv)}} biogeography based optimization (BBO), and {\em{(v)}} shuffled frog leaping algorithm (SFLA) to estimate the model parameters of the IB3DM. %Readers are referred to~\cite{} {\color{red} CITE MISSING} for more details on these optimization approaches.

The parameter is computed for two different PV technologies: monocrystalline (STM5-20/36) at 1000 \si{\watt\per\square\meter} at 33\si{\degree} and polycrystalline (Solartech SPM-020P-R) with 1000 \si{\watt\per\square\meter} at 45\si{\degree} with each panel having 36 cells in series.

  The IB3DM parameters computed for a monocrystalline panel for the SDM, DDM, CD3DM, and IB3DM with the five meta-heuristic algorithms is shown in Tab.~\ref{tab:Montable}. The RMSE value ($\mathcal{J}$) are shown in  Tab.~\ref{tab:Montable}.  One can observe that IB3DM offers better accuracy than existing models as evinced by their low RMSE values. In addition, fire-fly algorithm provides better estimates of the I-V curves.
  %In addition, the model parameters computed by the firefly algorithm provide a better approximation of I-V curves than other methods.

%The execution times for the different algorithms for parameter estimation for monocrystalline panel is shown in Tab.~\ref{tab:Exec}.  While firefly algorithm provided optimal parameters, TLBOs' computation time was less. The execution time for computing the model parameters for polycrystalline panel is shown in Tab.~\ref{tab:Exec1} which is similar to monocrystalline panel with TLBO showing excellent computation performance with decent accuracy, whereas the firefly had the least RMSE with high computation time.

	\begin{figure}[h]
		\centering
		%\subfigure[\label{1a}]{\includegraphics[width=0.45\textwidth]{7-1.pdf}}
		%\hspace{0.05\textwidth}
		\includegraphics[height=3.5cm, width=8cm]{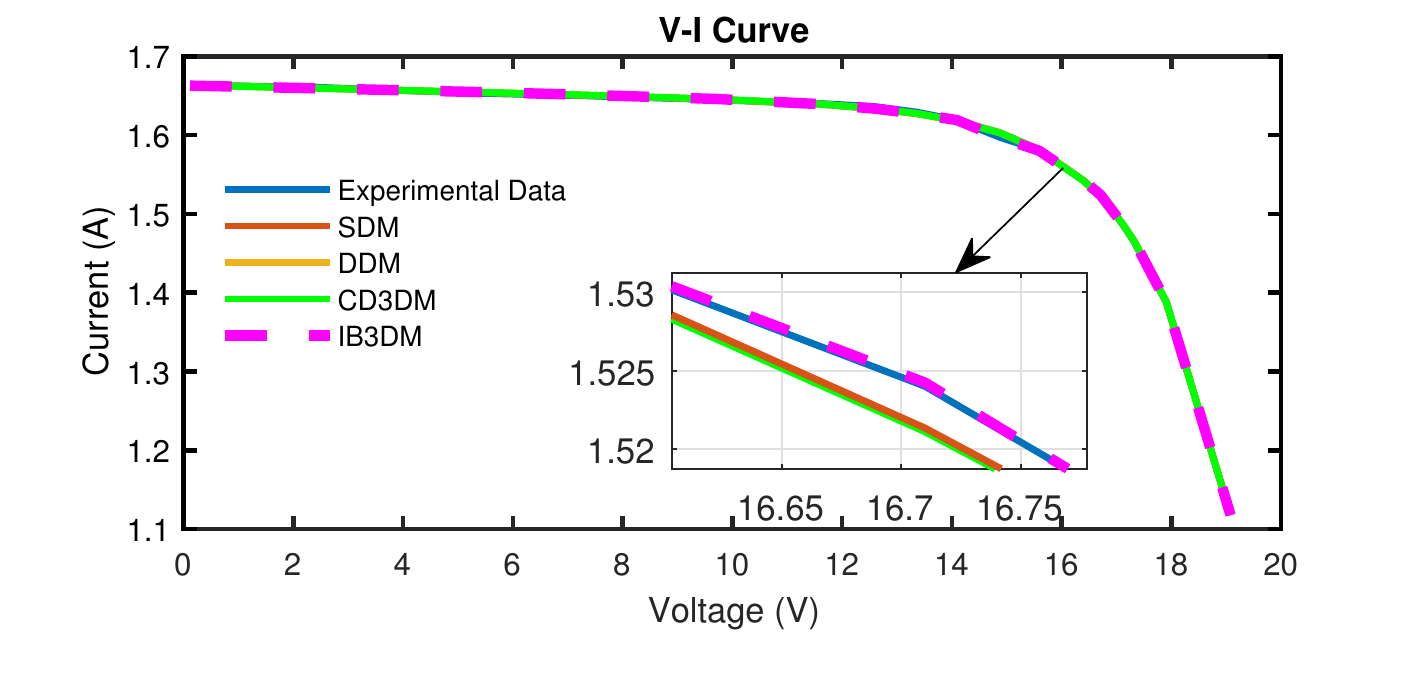}
			\hspace{0.05cm}	
				\includegraphics[height=3.5cm, width=8cm]{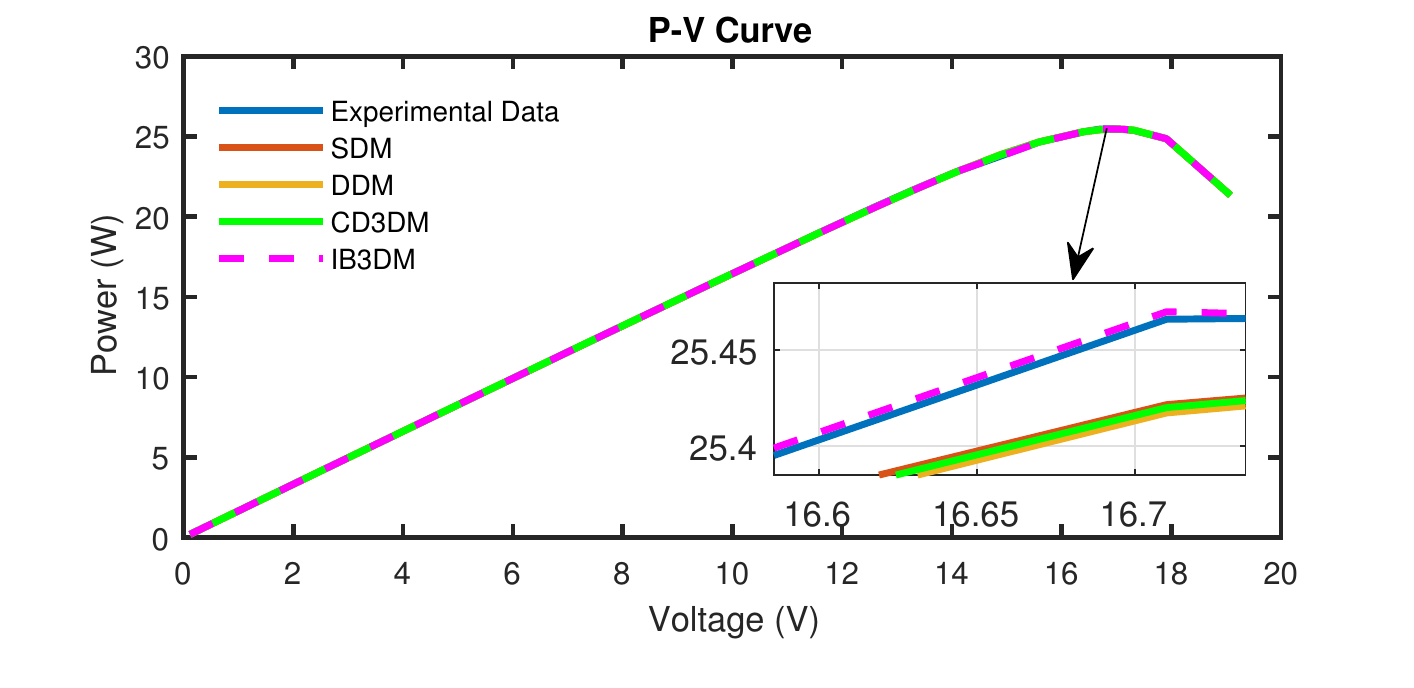}
		
		\caption{(a) V-I Curve of the SDM, DDM, C3DM versus IB3DM, (b) P-V Curve of SDM, DDM, C3DM, and IB3DM (Monocrystalline).}
		\label{fig:78}
	\end{figure}
	
		\begin{figure}
			\centering
				\includegraphics[height=3.8cm, width=8cm]{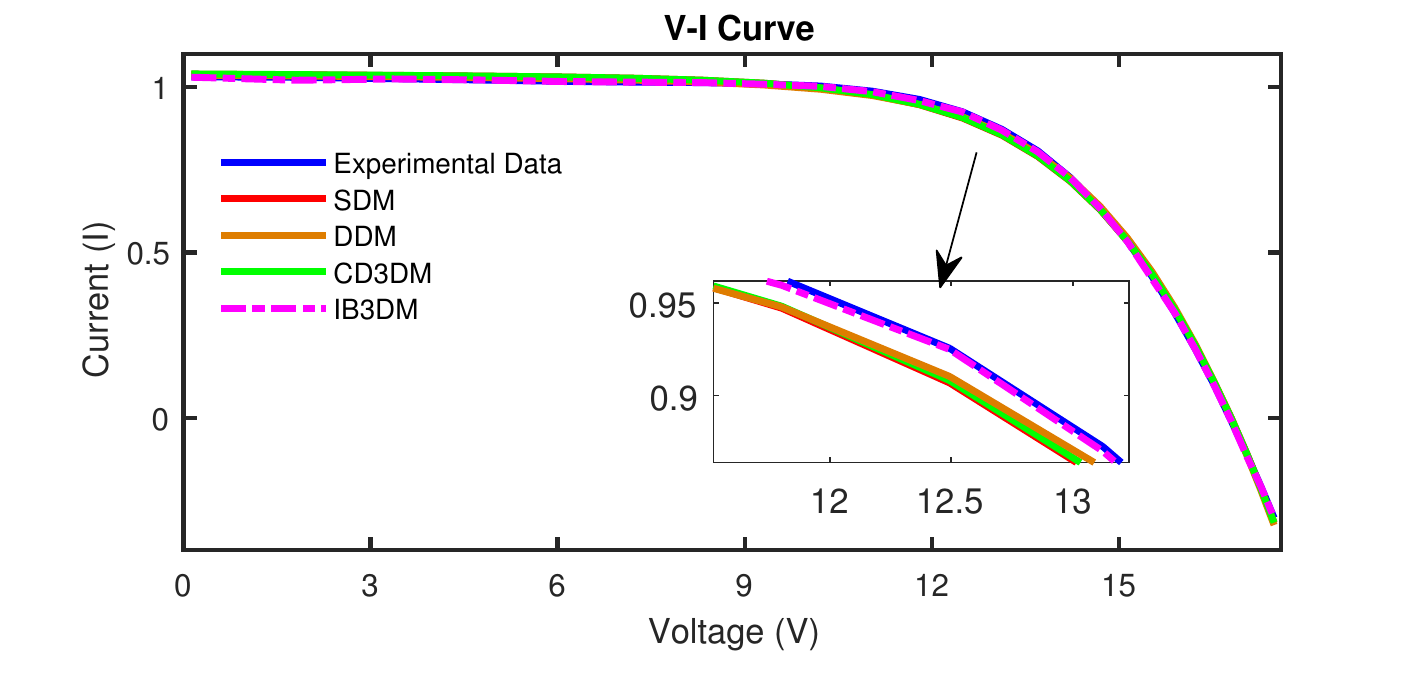}
			\hspace{0.05cm}	
				\includegraphics[height=4cm, width=8cm]{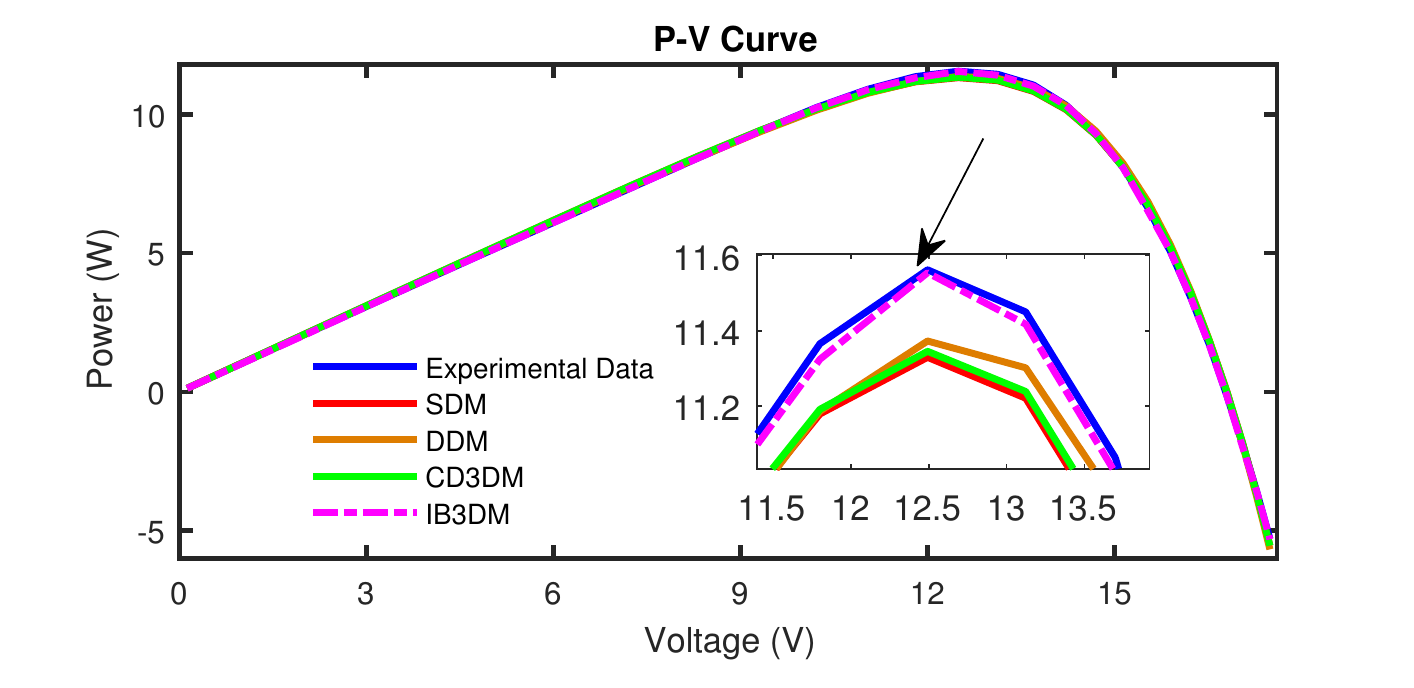}
		\caption{(a) V-I Curve of the SDM, DDM, C3DM versus IB3DM, (b) P-V Curve of SDM, DDM, C3DM, and IB3DM (Polycrystalline PV panel)}
		\label{fig:910}
	\end{figure}

\begin{table*}[h]
		\centering
		\caption{Comparison of meta-heuristic algorithms for computing the model parameters of the Monocrystalline panels}
		\label{tab:Montable}
		\begin{tabular}{| c| c |c |c |c |c| c| c |c| c |c| c| c |c| c| c| c| c| c| c| c| c| c |}
			\hline
			\textbf{Model} &\textbf{Algorithm} &\textbf{$I_{P}(A)$} &\textbf{$I_{01}(A)$} &\textbf{$I_{02}(A)$} &\textbf{$I_{03}(A)$} &\textbf{$n_{1}$} &\textbf{$n_{2}$} &\textbf{$n_{3}$}&\textbf{$R_{s}(\Omega)$} &\textbf{$R_{sh}(\Omega)$}&\textbf{$K$}&\textbf{$RMSE$} \\
			\hline
			
			\textbf{}&
			\textbf{Firefly}	&	1.6644	&	1.55E-06	&	-	&	-	&	1.94	&	-	&	-	&	0.176	&	752.29		&-&	0.012245\\
			\textbf{}&\textbf{PSO}	&	1.6626	&	2.87E-06	&	-	&	-	&	1.87	&	-	&	-	&	0.3917	&	599.78&	-&0.060392	\\
			\textbf{SDM}&\textbf{TLBO}	&	1.6634	&	2.86E-06	&	-	&	-	&	1.67	&	-	&	-	&	0.4255	&	598.55	&-	&0.042516\\
			\textbf{}&\textbf{BBO}	&	1.6605	&	9.08E-07	&	-	&	-	&	1.99	&	-	&	-	&	0.6372	&	642.66	&-&	0.065356	\\
			\textbf{}&\textbf{SFLA}	&	1.663	&	6.02E-06	&	-	&	-	&	1.96	&	-	&	-	&	0.2385	&	690.27		&-&	0.083196\\
			\hline
		\textbf{}&\textbf{Firefly}	&	1.6645	&	1.71E-06	&	3.01E-12	&	-	&	1.85	&	1.72	&	-	&	0.2396	&	739.49 &-&	0.010945	\\
			\textbf{}&\textbf{PSO}	&	1.7029	&	3.08E-05	&	6.24E-05	&	-	&	1.64	&	1.55	&	-	&	0.1263	&	606.28	&-&	0.052871\\
			\textbf{DDM}&\textbf{TLBO}	&	1.6638	&	1.14E-10	&	5.21E-06	&	-	&	1.52	&	1.91	&	-	&	0.5215	&	695.23 &-&	0.042925	\\
			\textbf{}&\textbf{BBO}	&	1.7029	&	3.08E-05	&	6.24E-05	&	-	&	1.94	&	1.55	&	-	&	0.3263	&	606.28 	&-&	0.085258	\\
			\textbf{}&\textbf{SFLA}	&	1.6613	&	5.66E-06	&	2.24E-08	&	-	&	1.76	&	1.89	&	-	&	0.2199	&	673.52	&-&	0.092876\\
			
			\hline
		\textbf{}&\textbf{Firefly}	&	1.6645	&	2.71E-06	&	3.01E-12	&	1.08E-05	&	1.72	&	1.57	&	1.47	&	0.296	&	739.49 &0.0092&	0.004852	\\
			\textbf{}&\textbf{PSO}	&	1.7029	&	1.08E-05	&	2.24E-05	&	1.72E-05	&	1.52	&	1.22	&	1.42	&	0.2263	&	656.28	&0.0272&	0.018471\\
			\textbf{CD3DM}&\textbf{TLBO}	&	1.6638	&	2.24E-10	&	3.31E-06	& 4.24E-08	&	1.49	&	1.19	&	1.24	&	0.3215	&	595.23 &0.0231&	0.009229	\\
			\textbf{}&\textbf{BBO}	&	1.7029	&	1.98E-05	&	2.41E-05	&	3.23E-08	&	1.76	&	1.28	&	1.62	&	0.2363	&	596.28 	&0.0185&	0.024528	\\
			\textbf{}&\textbf{SFLA}	&	1.6613	&	2.66E-06	&	5.24E-08	&	4.76E-06	&	1.62	&	1.24	&	1.32	&	0.2699	&	473.52	&0.0289&	0.050476\\
				
			\hline
			
				\textbf{}&\textbf{Firefly}	&	1.6633	&	2.93E-06	&	5.10E-15	&	1.54E-07	&	1.35	&	1.46	&	1.24	&	0.0917	&	804.43	&-&	{0.005463}\\
			\textbf{}&\textbf{PSO}	&	1.7133	&	6.88E-04	&	1.80E-10	&	1.63E-09	&	1.02	&	1.09	&	1.14	&	0.3618	&	477.24&-&	{0.007824}		\\
			\textbf{IB3DM}&\textbf{TLBO}	&	1.6622	&	1.89E-08	&	8.67E-08	&	1.19E-05	&	1.08	&	1.06	&	1.15	&	0.3917	&	761.51	&-&	{0.005936}
				\\
			\textbf{}&\textbf{BBO}	&	1.6683	&	8.52E-06	&	4.13E-06	&	2.85E-04	&	1.03	&	1.12	&	1.39	&	0.2511	&	570.46	&-&	{0.008193}	\\
			\textbf{}&\textbf{SFLA}	&1.6683	&	8.52E-06	&	4.13E-06	&	2.85E-04	&	1.13	&	1.18	&	1.29	&	0.3511	&	570.46	&-&	{0.008262}	\\
			\hline
		\end{tabular}
	\end{table*}

	\begin{table*}[!h]
		\centering
		\caption{ Comparison of meta-heuristic algorithms for computing the model parameters of the Polycrystalline panel}
		\label{tab:PolyTab}
		\begin{tabular}{| c| c |c |c |c |c| c| c |c| c |c| c| c |c| c| c| c| c| c| c| c| c| c |}
			\hline
			\textbf{Model} &\textbf{Algorithm} &\textbf{$I_{P}(A)$} &\textbf{$I_{01}(A)$} &\textbf{$I_{02}(A)$} &\textbf{$I_{03}(A)$} &\textbf{$n_{1}$} &\textbf{$n_{2}$} &\textbf{$n_{3}$}&\textbf{$R_{s}(\Omega)$} &\textbf{$R_{sh}(\Omega)$} &\textbf{$K$}&\textbf{$RMSE$}\\
			\hline	
			\textbf{}	&	\textbf{Firefly}	&	1.047	&	4.43E-05	&	-	&	-	&	1.85	&	-	&	-	&	1.7763	&	612.53	&-&	0.030324\\
			\textbf{}	&	\textbf{PSO}	&	1.0458	&	5.86E-05	&	-	&	-	&	1.96	&	-	&	-	&	1.7286	&	680.97	&-&	0.549642\\
			\textbf{SDM}	&	\textbf{TLBO}	&	1.0345	&	1.91E-05	&	-	&	-	&	1.86	&	-	&	-	&	1.9521	&	684.96	&-&	0.495359\\
			\textbf{}	&	\textbf{BBO}	&	1.0343	&	1.43E-04	&	-	&	-	&	1.97	&	-	&	-	&	1.4589	&	653.48	&-&	0.576291\\
			\textbf{}	&	\textbf{SFLA}	&	1.0466	&	1.31E-04	&	-	&	-	&	1.98	&	-	&	-	&	1.5583	&	699.79		&-&	0.762183\\
			
			\hline
			\textbf{}	&	\textbf{Firefly}	&	1.0435	&	5.15E-05	&	6.03E-13	&	-	&	1.83	&	1.65	&	-	&	1.7621	&	611.84	&-&	0.027368\\
			\textbf{}	&	\textbf{PSO}	&	1.0676	&	7.88E-05	&	6.03E-08	&	-	&	1.91	&	1.71	&	-	&	1.7272	&	591.83	&-&	0.389554\\
			\textbf{DDM}	&	\textbf{TLBO}	&	1.0323	&	2.14E-05	&	5.37E-06	&	-	&	1.79	&	1.69	&	-	&	1.1532	&	692.98	&-&	0.069885\\
			\textbf{}	&	\textbf{BBO}	&	1.0881	&	1.55E-05	&	6.62E-04	&	-	&	1.93	&	1.79	&	-	&	1.0243	&	641.61	&-&	0.401265\\
			\textbf{}	&	\textbf{SFLA}	&	1.0971	&	1.67E-04	&	3.53E-05	&	-	&	1.95	&	1.82	&	-	&	1.0004	&	358.15	&-&	0.495239\\
			\hline
			\textbf{}&\textbf{Firefly}	&	1.0745	&	1.62E-06	&	1.92E-12	&	3.28E-05	&	1.62	&	1.48	&	1.28	&	1.0235	&	798.25 &0.0052&	0.011619	\\
			\textbf{}&\textbf{PSO}	&	1.0929	&	6.54E-05	&	1.54E-05	&	1.92E-05	&	1.85	&	1.62	&	1.59	&	1.0563	&	696.28	&0.0237&	0.387132\\
			\textbf{CD3DM}&\textbf{TLBO}	&	1.6538	&	1.52E-10	&	2.26E-06	& 5.42E-08	&	1.68	&	1.34	&	1.45	&	1.5235	&	495.23 &0.0259&	0.065293	\\
			\textbf{}&\textbf{BBO}	&	1.0029	&	1.98E-05	&	2.41E-05	&	3.23E-08	&	1.89	&	1.67	&	1.62	&	1.0063	&	606.28 	&0.0262&	0.537698	\\
			\textbf{}&\textbf{SFLA}	&	1.0513	&	3.66E-08	&	3.24E-06	&	3.76E-09	&	1.91	&	1.73	&	1.71	&	1.0039	&	623.52	&0.0325&	0.553676\\

			\hline
		
		\textbf{}	&	\textbf{Firefly}	&	1.047	&	2.27E-10	&	2.39E-04	&	2.03E-12	&	1.42	&	1.37	&	1.06	&	1.0954	&	729.53&-	&	{0.004856}
				\\
			\textbf{}	&	\textbf{PSO}	&	1.049	&	8.33E-06	&	4.64E-04	&	3.70E-12	&	1.66	&	1.54	&	1.23	&	1.9449	&	693.99	&-	&	{0.006806}	\\
			\textbf{IB3DM}	&	\textbf{TLBO}	&	1.042	&	3.32E-06	&	1.06E-04	&	1.17E-05	&	1.53	&	1.45	&	1.03	&	1.2107	&	671.01		&-	&	{0.005077}\\
			\textbf{}	&	\textbf{BBO}	&	1.039	&	1.43E-05	&	4.11E-05	&	4.67E-05	&	1.72	&	1.62	&	1.35	&	1.5124	&	697.74		&-	&	{0.007493}\\
			\textbf{}	&	\textbf{SFLA}	&	1.045	&	2.63E-05	&	2.92E-07	&	2.26E-05	&	1.79	&	1.67	&	1.49	&	1.9119	&	691.77		&	-	&{0.007994}\\
			
			\hline
		\end{tabular}
	\end{table*}
	
A comparison of the I-V and P-V curves with model parameters computed by the firefly algorithm for the different PV models: SDM, DDM, CD3DM, and IB3DM is shown in Fig.~\ref{fig:78}. The accuracy provided by IB3DM at MPP is also shown in the zoomed portion. Similarly, accuracy is also high at low and high irradiance conditions as well.

 The IB3DM model parameters and its comparison with other PV models for poly-crystalline panel (SPM-020P-R) is shown in Tab.~\ref{tab:PolyTab}. The low RMSE values are indicative of the accuracy provided by IB3DM. Further, firefly algorithm provides the best model parameters compared with other meta-heuristic techniques.  Similarly, the I-V and P-V curves for the different diode model is shown in Fig.~\ref{fig:910}. This results shows the model accuracy across irradiance levels. These results demonstrate the IB3DMs' ability to provide model accuracy across different irradiance levels and PV technologies.

%One can see that the IB3DM provides better accuracy than existing models in the literature. Moreover, the model provides more accurate P-V curves at low irradiance conditions as well. These results indicate the capability of IB3DM to have better accuracy with respect to changes in PV technologies and existing PV models.

\subsection{Explainable Incipient Fault Detection}
Having computed the model parameters, the next step is to fuse IB3DM with XAI to implement the XFDDS. First, we show the ability of IB3DM to detect faults and then extensions to providing explanations are presented. Two studies used to illustrate XFDDS capabilities:

\begin{enumerate}
    \item[{\em{(i)}}] Single-cell in a module is partially shaded in a PV panel consisting of 36 cells in series;
    \item[{\em{(ii)}}] Most cells in a array are partially shaded (8 among 36) in different proportions (10-90\%).
\end{enumerate}
These two cases cover most scenarios envisaged during the PV panel operations and the incipient fault was created artificially for the study. %Furthermore, to generate the partial shading faults, fine paper with different sizes are used to cover the panel.

\subsection{Case Study 1}
\begin{figure}
\centering
%	\includegraphics[height=4cm, width=8cm]{polyvi1.pdf}
		%	\hspace{0.05cm}	
			%	\includegraphics[height=4cm, width=8cm]{polypv1.pdf}
%	\hspace{0.05cm}
	\includegraphics[width=0.35\textwidth]{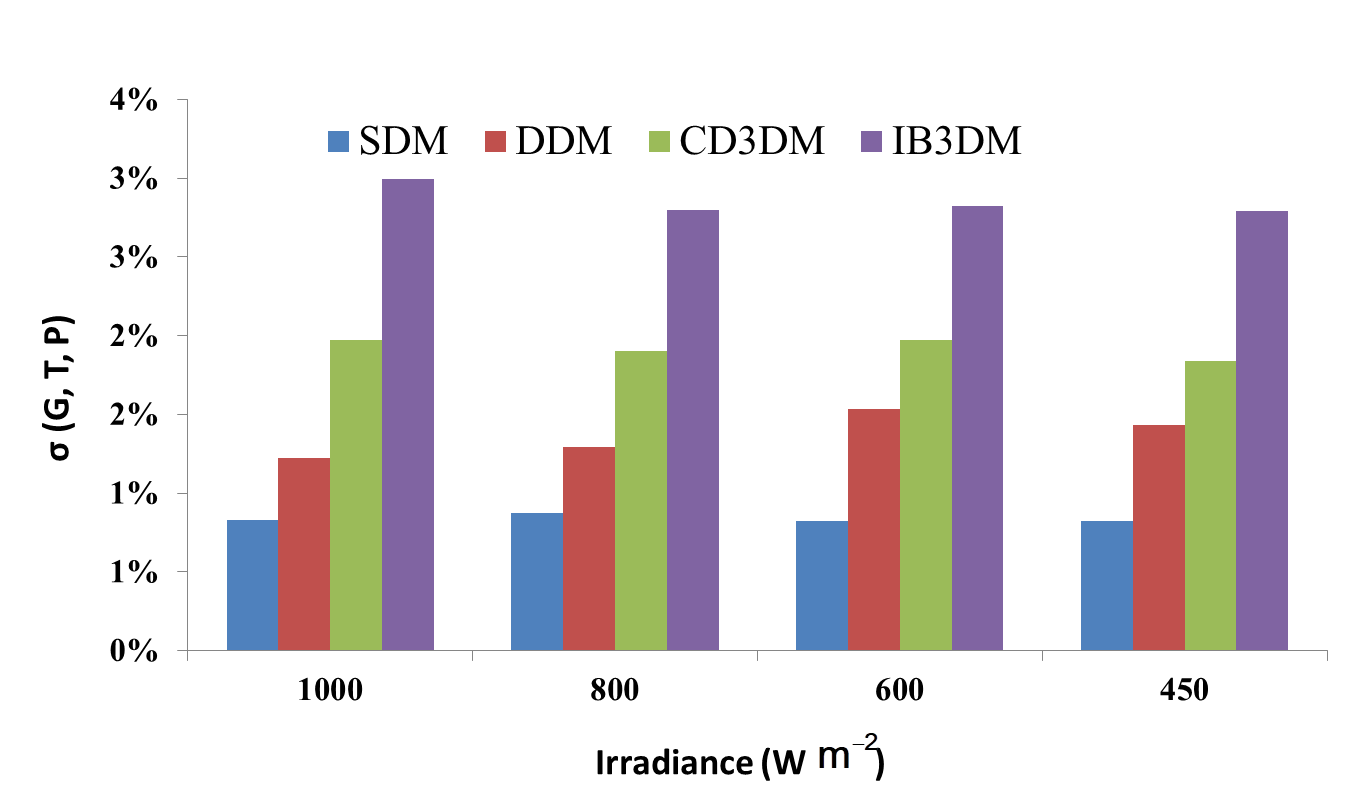}
	%\caption{(a).V-I Curve for single cell shaded., (b) P-V Curve for  single cell shaded,(c).Fault signature for different irradiance.}
	%\caption{(a) V-I Curve for different irradiance, and (b) P-V Curve for  different irradiance (polycrystalline).}
	\caption{Fault signature for different irradiance.}
\label{fig:28}
	\end{figure}

\begin{figure}
	\centering
	\includegraphics[height=3.0cm, width=8cm]{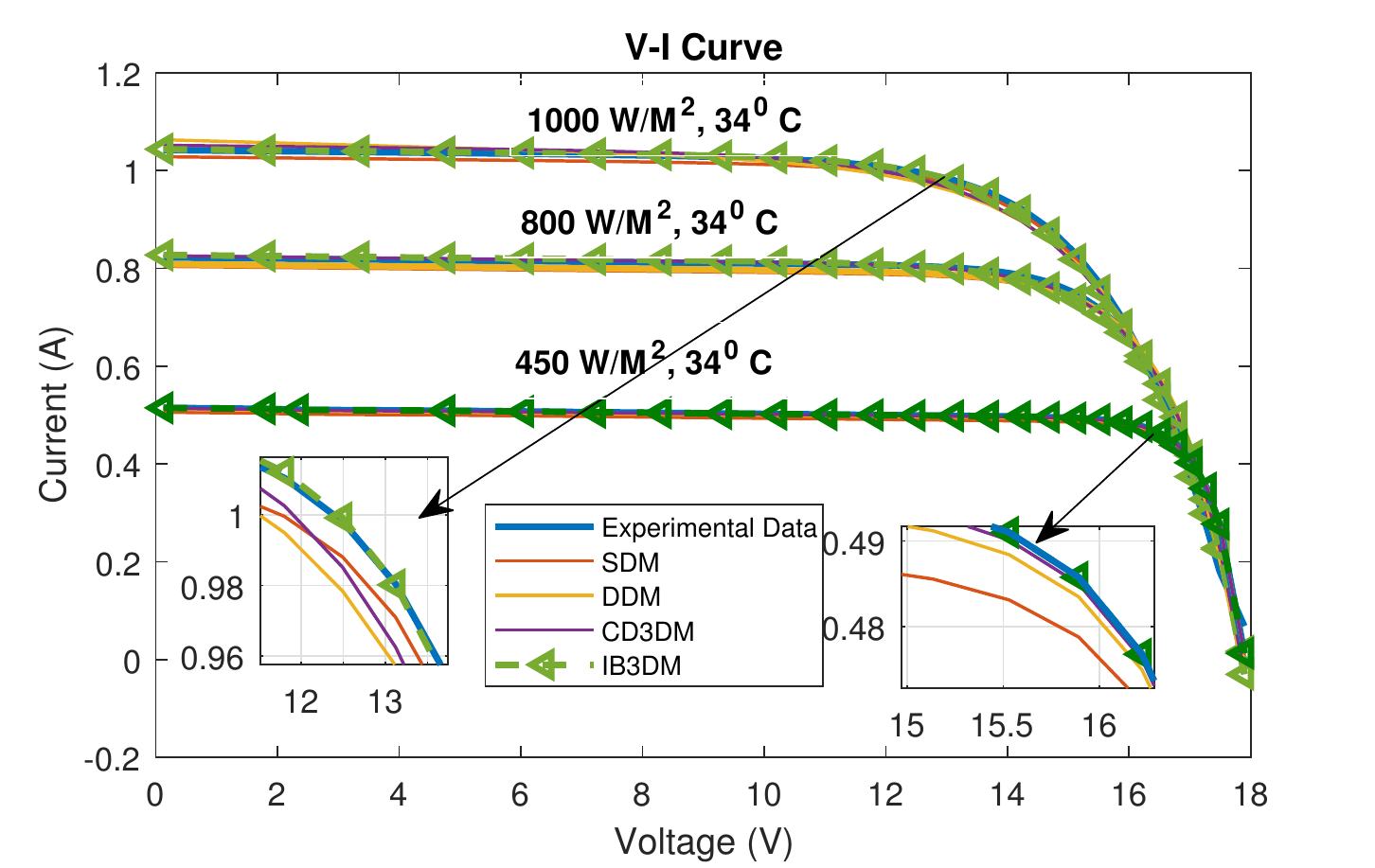}
			\hspace{0.05cm}	
				\includegraphics[height=3.0cm, width=8cm]{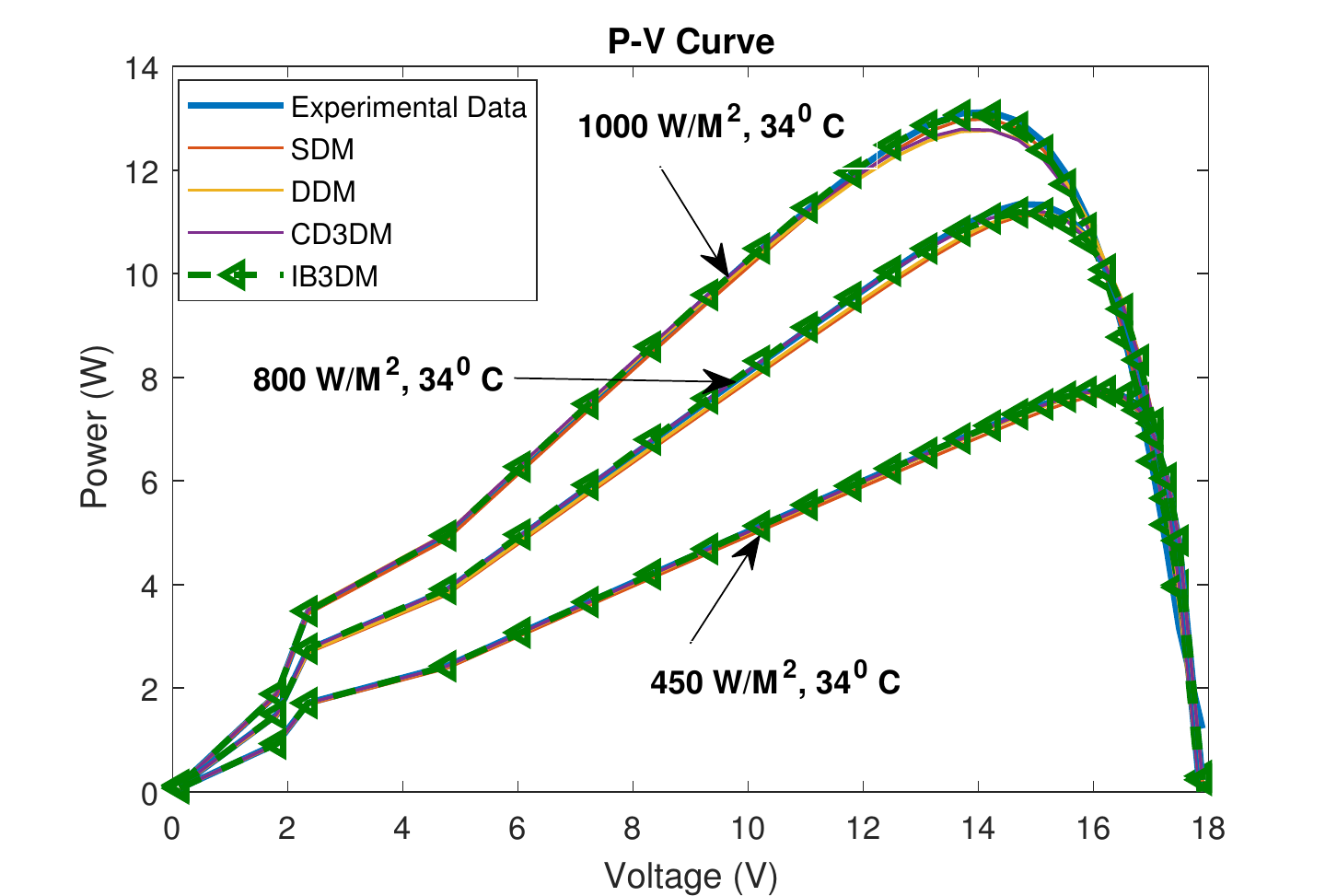}
	\hspace{0.05cm}
	\caption{(a) V-I Curve for different irradiance, and (b) P-V Curve for  different irradiance (polycrystalline).}
	\label{fig:25}
\end{figure}

This study considers  fault in  polycrystalline panel (Solartech SPM-020P-R) using IB3DM. To illustrate IB3DM ability to detect fault, the V-I and P-V curves are computed at different irradiance levels (0-1000 \si{\watt\per\square\meter}, 800 \si{\watt\per\square\meter}, and 450 \si{\watt\per\square\meter}). The I-V and P-V curves are shown in  Fig.\ref{fig:25} (a) and (b), respectively. One can see that the IB3DM predictions exactly coincide with experimental V-I and P-V curves, whereas model errors are high with other models.  Furthermore, slope of the PV curves at lower voltages are indicative of the incipient fault and this is illustrated from fault signatures at low irradiance conditions (see, Fig.~\ref{fig:28}). Using this fault-signature metric, the XFDDS could avoids false alarms (false positives). This results illustrates the ability of IB3DM to detect partial shading in single cell and avoid false alarms raised by XGBoost classifier. Moreover, detecting incipient faults is challenging with IB3DM alone.

\subsection{Case Study 2}
\begin{figure}[h]
\centering
\includegraphics[height=4.8cm, width=8cm]{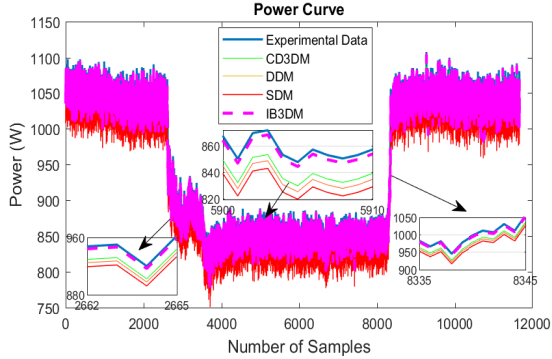}
\caption{Power curve for Partial shading condition for PV panel}
\label{fig:02}
\end{figure}

The IB3DMs' ability to detect faults in \emph{monocrystalline array} with partial shading of different magnitudes across PV panel is illustrated. Variations in power curves with the SDM, DDM, CD3DM, and IB3DM for irradiance levels (0-1000 \si{\watt\per\square\meter}) are considered. Due to partial shading, the voltage reduction is not severe, whereas the current reduction is quite high which gets reflected in the power curve as well. This is illustrated in Fig.~\ref{fig:02} and is indicative of a fault. This is observed in fault-signature metric as well. These findings when combined with XGBoost based classifier could reduce mis-classification and false alarms.

%This is indicative of the incipient faults. The fault signature metric defined captures the presence of an incipient fault and triggers the XAI application.

There are two shortcomings with just IB3DM based detection. It requires enough significant sample averages to detect incipient faults which is seldom possible due to their intermittent nature. Next, they cannot identify fault types nor provide explanations. 

%The results presented shows the IB3DM parameter computation and its ability to capture faults even at low irradiance conditions and for well developed faults which show prolonged fault-signature. However, incipient faults are intermittent and could last for a few samples. Therefore, their detection becomes difficult. To detect incipient faults and provide explanations, the XFDDS uses the XAI application.

\subsection{XFDDS for Incipient Faults}
The XFDDS is implemented on two different XGBoost implementations, binary and multi-class classification. The XGBoost leverages data and averts false alarms by using IB3DM fault-signature metric. In this study, the monocrystalline panel string consisting of 56 panels at the International Research Center, Kalasalingam University, is used as the pilot to demonstrate the XFDDS. In binary classification, the XGBoost classifies whether a given test sample is faulty/healthy. This is compared with the fault-signature metric outcomes. Using a threshold on FSM and based on XGBoost classifier outcomes, explanations are triggered.

\begin{figure}[h]
	\centering
	\includegraphics[height=2.5cm, width=8cm]{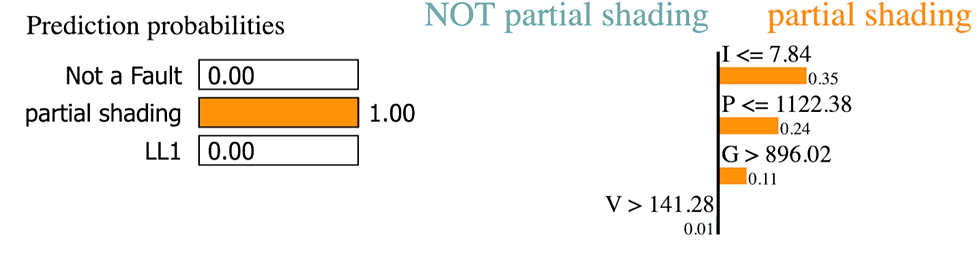}
			\hspace{0.05cm}	
	\caption{Partial shading fault}
	\label{fig:partial}
\end{figure}

 In our study, the partial shading fault was created artificially by covering panels with papers. A sampling time of 1~\si{\second} and the fault is created at 2400 samples for a duration of 32 \rm{minutes}. These samples are passed to the IB3DM, which detects the presence of incipient fault through its fault signature. This triggers the XAI application to generate explanations. The explanations for partial shading faults are shown in Fig.~\ref{fig:partial}. The explanations are very intuitive for detecting incipient partial shading fault. As the explanation shows that the irradiance (G $\ge$ 896) and the voltage (V $\ge$ 141.29 \si{\volt}) , but the current ($\le$ 9.77 \si{\ampere} ) and the power magnitudes ($\le$ 1354 \si{\watt}) are low. This is indicative of a partial shading fault for the field engineer.

\begin{figure}[h]
    \centering
    \includegraphics[scale = 0.30]{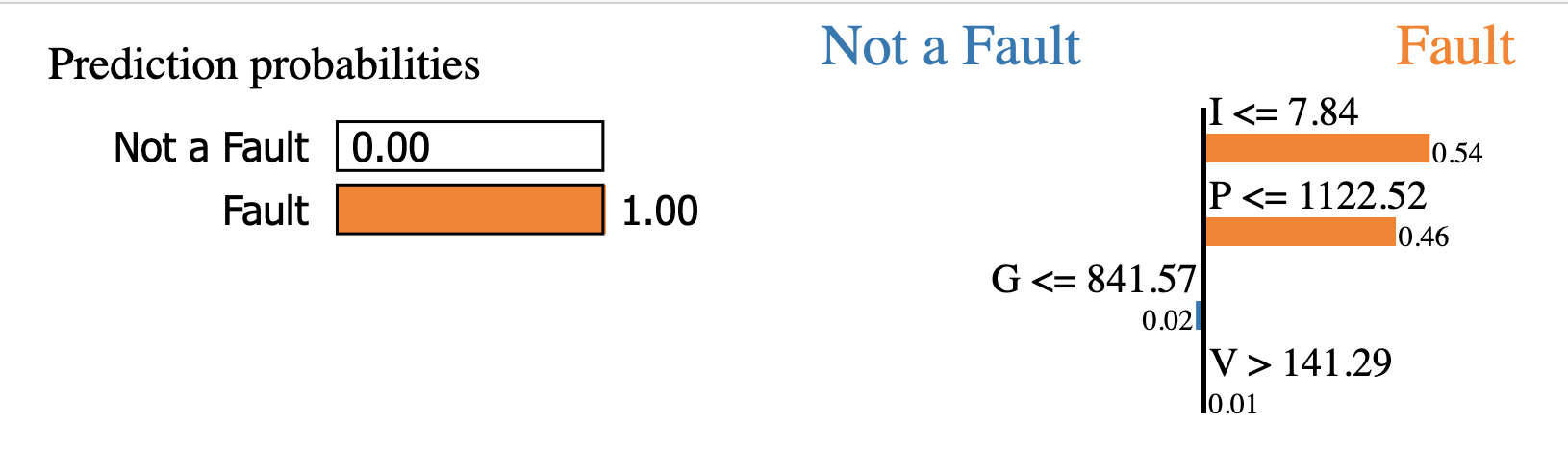}
    \caption{Line-to-Line Fault}
    \label{fig:LL}
\end{figure}

Similarly, the LL fault was created by shorting out the lines within a single string for a short duration. The IB3DM generates fault-signature of the power curve, which triggered the explanations that are shown in Fig.~\ref{fig:LL} which is explaining that current values are less than 7.44\si{\ampere} and power values lesser than 1112.52 \si{\watt} while the voltage is greater than 141.29 \si{\volt} helps the XFDDS decide that there is a LL fault.

\begin{figure}
    \centering
    \includegraphics[scale = 0.30]{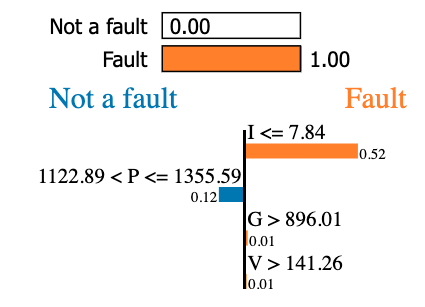}
    \caption{Line-to-Line Fault Explanation}
    \label{fig:LL1}
\end{figure}

Explanations on a sample which was false positive detected by XGBoost and explanations provided by the XAI is shown in Fig.\ref{fig:LL1}. Here the decision was based on the current value, which was less than 7.84 \si{\ampere}. However, the power level greater than 1112.89 \si{\watt} shows that this is not a fault. This way, false positives could be avoided by explanations, and this denotes an intermittent behaviour of the load to the field engineer.

\begin{figure}
	\centering
	\includegraphics[height=2cm, width=8cm]{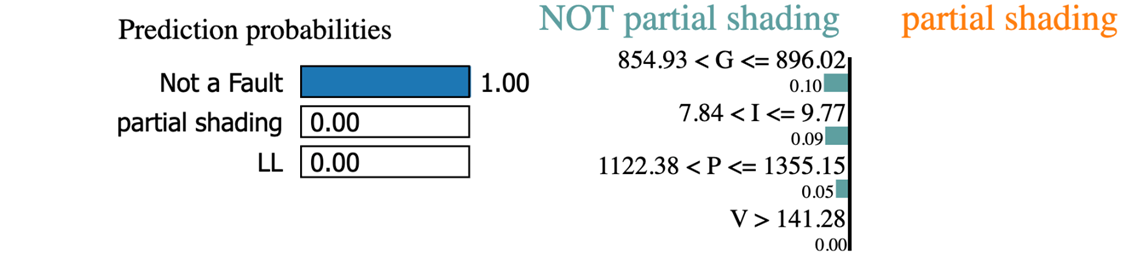}
	\caption{Explanations for LL-fault Multi-Label Classification}
	\label{fig:MLLL}
		\end{figure}

Next, the XFDDS was applied to multi-class classification problem. In this case, the fault-labels were given as input as well. The IB3DM was used to trigger explanations using thresholds on fault-signature metric. The explanations for the incipient LL fault is shown in Fig.~\ref{fig:MLLL}. In this case the XGBoost identifies the LL fault and the causes are illustrated by current ( less than 7.84 \si{\ampere}) and power (less than 1122.38 \si{\watt}). 
\begin{figure}[h]
	\centering
	 \includegraphics[scale=0.30]{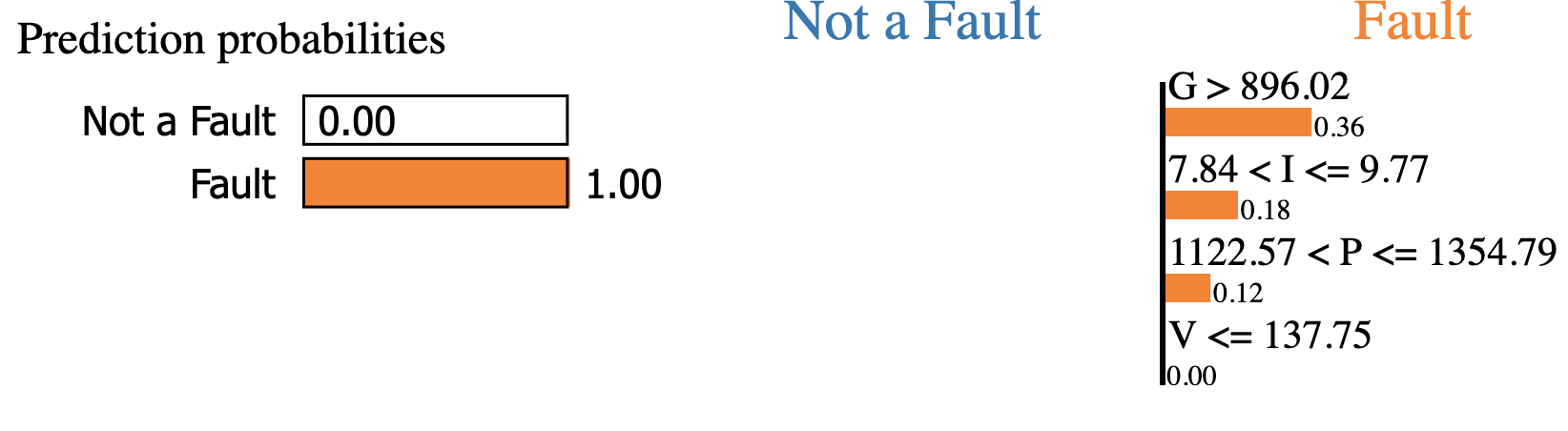}
	\caption{Explanations for Partial Shading Faults}
	\label{fig:PSLL}
		\end{figure}

The explanations for partial shading faults are shown in Fig.~\ref{fig:PSLL}. The explanations are very intuitive as the current and power are less than a threshold, whereas the irradiance is greater than 896 \si{\watt\per\square\meter}. The fault could be easily understood by the field technicians. Furthermore, the threshold values are explained to the field engineers on current, power, irradiance, and voltage. This result also shows that the explanations and the thresholds are oblivious to the classification label, i.e., binary or multi-class.

\vspace{-0.3cm}

	\section{Conclusions}
	This paper presented an explainable fault-detection and diagnosis system (XFDDS) for detecting incipient faults in PV panels. Its main components were: irradiance based three diode model and eXplainable artificial intelligence (XAI) application. The IB3DM used irradiance and temperature to compute its model parameters, thereby increasing its accuracy even at low irradiation conditions. The model parameters were computed solving a non-convex constrained optimization problem with five different meta-heuristic approaches. Our results demonstrated the model fidelity of the IB3DM even at low irradiance conditions. To exploit the aggregated data from PV panels, an extreme gradient boosting (XGBoost) based classifier was used and it had two shortcomings: false alarms and lack of explainability. False alarms were reduced by combining IB3DM with XGBoost classifier by proposing a fault-signature metric (FSM). Explanations were provided by extending the XGBoost with local interpretable model-agnostic explanations (LIME) in the XAI application. The explanations are quite useful in identifying faults and planning maintenance operations. The proposed XFDDS implementation, deployment and demonstration on multiple PV technologies was used to illustrate the capabilities. Extending XFDDS to include multiple faults and translating the approach to an edge are the future course of this investigation.

 % Finally, the XFDDS was demonstrated on different PV technologies, and the explainability aspect was illustrated. Extending XFDDS to other faults and scaling the solution are the prospects of the work.

%	\vspace{-0.1cm}
%	\section*{Acknowledgments}
%	This project is supported by the project Resilient and Optimal-micro-
%energy grid (ROME): Funded by Department of Science and Tech-
%nology and Research Council of Norway, Norway through Resilient
%and Optimal Micro-Energy grid INT/NOR/RCN/ICT/P-05/2018.
	\bibliographystyle{IEEEtran}

%\bibliography{references}
\bibliography{references}		

\begin{thebibliography}{10}
\providecommand{\url}[1]{#1}
\csname url@samestyle\endcsname
\providecommand{\newblock}{\relax}
\providecommand{\bibinfo}[2]{#2}
\providecommand{\BIBentrySTDinterwordspacing}{\spaceskip=0pt\relax}
\providecommand{\BIBentryALTinterwordstretchfactor}{4}
\providecommand{\BIBentryALTinterwordspacing}{\spaceskip=\fontdimen2\font plus
\BIBentryALTinterwordstretchfactor\fontdimen3\font minus
  \fontdimen4\font\relax}
\providecommand{\BIBforeignlanguage}[2]{{%
\expandafter\ifx\csname l@#1\endcsname\relax
\typeout{** WARNING: IEEEtran.bst: No hyphenation pattern has been}%
\typeout{** loaded for the language `#1'. Using the pattern for}%
\typeout{** the default language instead.}%
\else
\language=\csname l@#1\endcsname
\fi
#2}}
\providecommand{\BIBdecl}{\relax}
\BIBdecl

\bibitem{rajesh2020design}
T.~Rajesh, K.~Tamilselvan, A.~Vijayalakshmi, C.~N. Kumar, and K.~A. Reddy,
  ``Design and implementation of an automatic solar tracking system for a
  monocrystalline silicon material panel using mppt algorithm,''
  \emph{Materials Today: Proceedings}, 2020.

\bibitem{karmakar2020detection}
B.~K. Karmakar and A.~K. Pradhan, ``Detection and classification of faults in
  solar pv array using thevenin equivalent resistance,'' \emph{IEEE Journal of
  Photovoltaics}, vol.~10, no.~2, pp. 644--654, 2020.

\bibitem{zhao2014graph}
Y.~Zhao, R.~Ball, J.~Mosesian, J.-F. de~Palma, and B.~Lehman, ``Graph-based
  semi-supervised learning for fault detection and classification in solar
  photovoltaic arrays,'' \emph{IEEE Transactions on Power Electronics},
  vol.~30, no.~5, pp. 2848--2858, 2014.

\bibitem{pillai2019comparative}
D.~S. Pillai, F.~Blaabjerg, and N.~Rajasekar, ``A comparative evaluation of
  advanced fault detection approaches for pv systems,'' \emph{IEEE Journal of
  Photovoltaics}, vol.~9, no.~2, pp. 513--527, 2019.

\bibitem{jin2019}
B.~Jin, D.~Li, S.~Srinivasan, S.-K. Ng, K.~Poolla, and
  A.~Sangiovanni-Vincentelli, ``Detecting and diagnosing incipient building
  faults using uncertainty information from deep neural networks,'' in
  \emph{2019 IEEE International Conference on Prognostics and Health Management
  (ICPHM)}.\hskip 1em plus 0.5em minus 0.4em\relax IEEE, 2019, pp. 1--8.

\bibitem{garoudja2017statistical}
E.~Garoudja, F.~Harrou, Y.~Sun, K.~Kara, A.~Chouder, and S.~Silvestre,
  ``Statistical fault detection in photovoltaic systems,'' \emph{Solar Energy},
  vol. 150, pp. 485--499, 2017.

\bibitem{mellit2018fault}
A.~Mellit, G.~M. Tina, and S.~A. Kalogirou, ``Fault detection and diagnosis
  methods for photovoltaic systems: A review,'' \emph{Renewable and Sustainable
  Energy Reviews}, vol.~91, pp. 1--17, 2018.

\bibitem{gao2015}
Z.~Gao, C.~Cecati, and S.~X. Ding, ``A survey of fault diagnosis and
  fault-tolerant techniques—part i: Fault diagnosis with model-based and
  signal-based approaches,'' \emph{IEEE Transactions on Industrial
  Electronics}, vol.~62, no.~6, pp. 3757--3767, 2015.

\bibitem{chaibi2019simple}
Y.~Chaibi, M.~Malvoni, A.~Chouder, M.~Boussetta, and M.~Salhi, ``Simple and
  efficient approach to detect and diagnose electrical faults and partial
  shading in photovoltaic systems,'' \emph{Energy Conversion and Management},
  vol. 196, pp. 330--343, 2019.

\bibitem{shongwe2015comparative}
S.~Shongwe and M.~Hanif, ``Comparative analysis of different single-diode pv
  modeling methods,'' \emph{IEEE Journal of photovoltaics}, vol.~5, no.~3, pp.
  938--946, 2015.

\bibitem{bradaschia2019parameter}
F.~Bradaschia, M.~C. Cavalcanti, A.~J. do~Nascimento, E.~A. da~Silva, and G.~M.
  de~Souza~Azevedo, ``Parameter identification for pv modules based on an
  environment-dependent double-diode model,'' \emph{IEEE Journal of
  Photovoltaics}, vol.~9, no.~5, pp. 1388--1397, 2019.

\bibitem{khanna2015three}
V.~Khanna, B.~Das, D.~Bisht, P.~Singh \emph{et~al.}, ``A three diode model for
  industrial solar cells and estimation of solar cell parameters using pso
  algorithm,'' \emph{Renewable Energy}, vol.~78, pp. 105--113, 2015.

\bibitem{qais2019identification}
M.~H. Qais, H.~M. Hasanien, and S.~Alghuwainem, ``Identification of electrical
  parameters for three-diode photovoltaic model using analytical and sunflower
  optimization algorithm,'' \emph{Applied Energy}, vol. 250, pp. 109--117,
  2019.

\bibitem{triki2018fault}
A.~Triki-Lahiani, A.~B.-B. Abdelghani, and I.~Slama-Belkhodja, ``Fault
  detection and monitoring systems for photovoltaic installations: A review,''
  \emph{Renewable and Sustainable Energy Reviews}, vol.~82, pp. 2680--2692,
  2018.

\bibitem{davarifar2013real}
M.~Davarifar, A.~Rabhi, A.~El-Hajjaji, and M.~Dahmane, ``Real-time model base
  fault diagnosis of pv panels using statistical signal processing,'' in
  \emph{2013 International Conference on Renewable Energy Research and
  Applications (ICRERA)}.\hskip 1em plus 0.5em minus 0.4em\relax IEEE, 2013,
  pp. 599--604.

\bibitem{fadhel2019pv}
S.~Fadhel, C.~Delpha, D.~Diallo, I.~Bahri, A.~Migan, M.~Trabelsi, and
  M.~Mimouni, ``Pv shading fault detection and classification based on iv curve
  using principal component analysis: Application to isolated pv system,''
  \emph{Solar Energy}, vol. 179, pp. 1--10, 2019.

\bibitem{ali2017real}
M.~H. Ali, A.~Rabhi, A.~El~Hajjaji, and G.~M. Tina, ``Real time fault detection
  in photovoltaic systems,'' \emph{Energy Procedia}, vol. 111, pp. 914--923,
  2017.

\bibitem{chen2017adaptive}
L.~Chen and X.~Wang, ``Adaptive fault localization in photovoltaic systems,''
  \emph{IEEE Transactions on Smart Grid}, vol.~9, no.~6, pp. 6752--6763, 2017.

\bibitem{yi2017line}
Z.~Yi and A.~H. Etemadi, ``Line-to-line fault detection for photovoltaic arrays
  based on multiresolution signal decomposition and two-stage support vector
  machine,'' \emph{IEEE Transactions on Industrial Electronics}, vol.~64,
  no.~11, pp. 8546--8556, 2017.

\bibitem{yi2016fault}
------, ``Fault detection for photovoltaic systems based on multi-resolution
  signal decomposition and fuzzy inference systems,'' \emph{IEEE Transactions
  on Smart Grid}, vol.~8, no.~3, pp. 1274--1283, 2016.

\bibitem{zhao2019artificial}
Y.~Zhao, T.~Li, X.~Zhang, and C.~Zhang, ``Artificial intelligence-based fault
  detection and diagnosis methods for building energy systems: Advantages,
  challenges and the future,'' \emph{Renewable and Sustainable Energy Reviews},
  vol. 109, pp. 85--101, 2019.

\bibitem{dhibi2020reduced}
K.~Dhibi, R.~Fezai, M.~Mansouri, M.~Trabelsi, A.~Kouadri, K.~Bouzara,
  H.~Nounou, and M.~Nounou, ``Reduced kernel random forest technique for fault
  detection and classification in grid-tied pv systems,'' \emph{IEEE Journal of
  Photovoltaics}, 2020.

\bibitem{zhao2020collaborative}
Y.~Zhao, D.~Li, T.~Lu, Q.~Lv, N.~Gu, and L.~Shang, ``Collaborative fault
  detection for large-scale photovoltaic systems,'' \emph{IEEE Transactions on
  Sustainable Energy}, 2020.

\bibitem{gan2019novel}
Y.~Gan, Z.~Chen, L.~Wu, C.~Long, S.~Cheng, and P.~Lin, ``A novel fault
  diagnosis method for pv arrays using extreme gradient boosting classifier,''
  2019.

\bibitem{ebrahimifakhar2020data}
A.~Ebrahimifakhar, A.~Kabirikopaei, and D.~Yuill, ``Data-driven fault detection
  and diagnosis for packaged rooftop units using statistical machine learning
  classification methods,'' \emph{Energy and Buildings}, vol. 225, p. 110318,
  2020.

\bibitem{li2020survey}
X.-H. Li, C.~C. Cao, Y.~Shi, W.~Bai, H.~Gao, L.~Qiu, C.~Wang, Y.~Gao, S.~Zhang,
  X.~Xue \emph{et~al.}, ``A survey of data-driven and knowledge-aware
  explainable ai,'' \emph{IEEE Transactions on Knowledge and Data Engineering},
  2020.

\bibitem{bramhall2020qlime}
S.~Bramhall, H.~Horn, M.~Tieu, and N.~Lohia, ``Qlime-a quadratic local
  interpretable model-agnostic explanation approach,'' \emph{SMU Data Science
  Review}, vol.~3, no.~1, p.~4, 2020.

\bibitem{torres2019regression}
A.~Torres-Barr{\'a}n, {\'A}.~Alonso, and J.~R. Dorronsoro, ``Regression tree
  ensembles for wind energy and solar radiation prediction,''
  \emph{Neurocomputing}, vol. 326, pp. 151--160, 2019.

\bibitem{chin2015cell}
V.~J. Chin, Z.~Salam, and K.~Ishaque, ``Cell modelling and model parameters
  estimation techniques for photovoltaic simulator application: A review,''
  \emph{Applied Energy}, vol. 154, pp. 500--519, 2015.

\bibitem{sun2020gradient}
R.~Sun, G.~Wang, W.~Zhang, L.-T. Hsu, and W.~Y. Ochieng, ``A gradient boosting
  decision tree based gps signal reception classification algorithm,''
  \emph{Applied Soft Computing}, vol.~86, p. 105942, 2020.

\bibitem{fan2018comparison}
J.~Fan, X.~Wang, L.~Wu, H.~Zhou, F.~Zhang, X.~Yu, X.~Lu, and Y.~Xiang,
  ``Comparison of support vector machine and extreme gradient boosting for
  predicting daily global solar radiation using temperature and precipitation
  in humid subtropical climates: A case study in china,'' \emph{Energy
  Conversion and Management}, vol. 164, pp. 102--111, 2018.

\bibitem{sapountzoglou2020fault}
N.~Sapountzoglou, J.~Lago, and B.~Raison, ``Fault diagnosis in low voltage
  smart distribution grids using gradient boosting trees,'' \emph{Electric
  Power Systems Research}, vol. 182, p. 106254, 2020.

\end{thebibliography}
	
\end{document}